\documentclass[conference]{IEEEtran}
\IEEEoverridecommandlockouts
\usepackage{cite}
\usepackage{amsmath,amssymb,amsfonts}
\usepackage{algorithmic}
\usepackage{graphicx}
\usepackage{textcomp, tabularray, setspace, subcaption, adjustbox, amsmath, amsfonts}
\usepackage{xcolor}
\usepackage{fancyhdr}

\usepackage{pgfgantt} % Gantt Chart
\usepackage{rotating}

\usepackage{graphicx} % Required for inserting images
\usepackage{wrapfig}
\usepackage{enumerate}
\usepackage{hyperref}
\usepackage{xurl}
\usepackage{tabularx}
\usepackage{url}
\usepackage{dcolumn}
\usepackage{listings}
\usepackage{algorithm}
\usepackage{amsmath}
\usepackage{verbatim}
\usepackage{float}% Required for Figures to float at the specified position
\usepackage{graphicx}

\title{Integrating Saliency Ranking and Reinforcement Learning for Enhanced Object Detection}

\author{
    \IEEEauthorblockN{Matthias Bartolo}
    \IEEEauthorblockA{\textit{Department of AI}  \\ \textit{University of Malta} \\
    \href{mailto:matthias.bartolo.21@um.edu.mt}{matthias.bartolo.21@um.edu.mt}}

    \and
    \IEEEauthorblockN{Dylan Seychell}
    \IEEEauthorblockA{\textit{Department of AI}  \\ \textit{University of Malta}  \\
    \href{mailto:dylan.seychell@um.edu.mt}{dylan.seychell@um.edu.mt}}

    \and
    \IEEEauthorblockN{Josef Bajada}
    \IEEEauthorblockA{\textit{Department of AI}  \\ \textit{University of Malta}  \\
    \href{mailto:josef.bajada@um.edu.mt}{josef.bajada@um.edu.mt}}
}

\date{24th July 2024}

\begin{document}

\maketitle
\thispagestyle{plain}%For page numbers
\pagestyle{plain}

\begin{abstract}

With the ever-growing variety of object detection approaches, this study explores a series of experiments that combine reinforcement learning (RL)-based visual attention methods with saliency ranking techniques to investigate transparent and sustainable solutions. By integrating saliency ranking for initial bounding box prediction and subsequently applying RL techniques to refine these predictions through a finite set of actions over multiple time steps, this study aims to enhance RL object detection accuracy.
Presented as a series of experiments, this research investigates the use of various image feature extraction methods and explores diverse Deep Q-Network (DQN) architectural variations for deep reinforcement learning-based localisation agent training. Additionally, we focus on optimising the detection pipeline at every step by prioritising lightweight and faster models, while also incorporating the capability to classify detected objects, a feature absent in previous RL approaches.
We show that by evaluating the performance of these trained agents using the Pascal VOC 2007 dataset, faster and more optimised models were developed. Notably, the best mean Average Precision (mAP) achieved in this study was 51.4, surpassing benchmarks set by RL-based single object detectors in the literature.

\end{abstract}

\section{Introduction}%
\label{chp:intro}
Object detection, within the area of computer vision, is a critical problem in recognising and localising objects inside an image or video \cite{deepRLinCV}. At first sight, the challenge appears simple: identify objects in visual scenes. However, the difficulty stems from the sheer variation of objects in terms of size, shape, orientation, occlusion, and lighting. Furthermore, the context in which these objects occur complicates the process even more, mandating models that can generalise across varied contexts and views.

While substantial progress has been made in advancing this field with technologies such as YOLOv10 \cite{yolov10}, which utilise various computer algorithms to overcome these issues, further experimentation is required to achieve human-like competency. To this end, grasping the intricacies of human visual perception \cite{human_perception, itti, itti_ior} and understanding how humans locate objects are crucial aspects of this effort.

In \cite{object_detection_philospy}, T. Boger and T. Ullman conducted several tests to investigate individuals' capacity to locate objects. Their findings show that, regardless of object realism, judgements based on physical reasoning, namely centre of mass, are consistently the most significant element in perceived object location. Successfully showing a dependence on physical qualities to determine object location. In a broader sense, this study provides an excellent justification for investigating a fresh paradigm in the modelling of spatial perception inside visual frameworks, such as human perception in object detection \cite{object_detection_philospy}.

Thus, this study aims to investigate whether the integration of reinforcement learning and saliency ranking techniques, which mimic human perception, can effectively address the problem of object detection. We also explore the individual components of reinforcement learning-based object detectors with the goal of developing faster and more optimised models, while also examining the accuracy of the proposed architecture in comparison to the literature. Moreover, this research investigates whether adopting a reinforcement learning framework facilitates more transparent detection by enabling monitoring of the object detection training process.

The first part of this paper consists of background information on methodologies employed by saliency ranking and reinforcement learning, as well as a comprehensive review of past reinforcement learning techniques for object detection. Chapter \ref{chp:methodology} provides a full overview of the solution architecture and its modules. This is followed by a series of experiments, and a thorough comparative analysis in Chapter \ref{chp:eval}. The paper summarises the findings in Chapter \ref{chp:conclusion} and discusses proposed directions for future work in Chapter \ref{chp:future_work}.

\section{Background}%
\label{chp:background}
This chapter presents background knowledge that readers must familiarise themselves with before proceeding to the subsequent chapters, which served as the foundation for this endeavour. A brief synopsis of feature and classification learning, saliency ranking, and reinforcement learning is provided.

\subsection{Saliency Ranking}
\label{sec:ranking}
\begin{figure}[ht]
    \centering
    \includegraphics[width=0.9\linewidth]{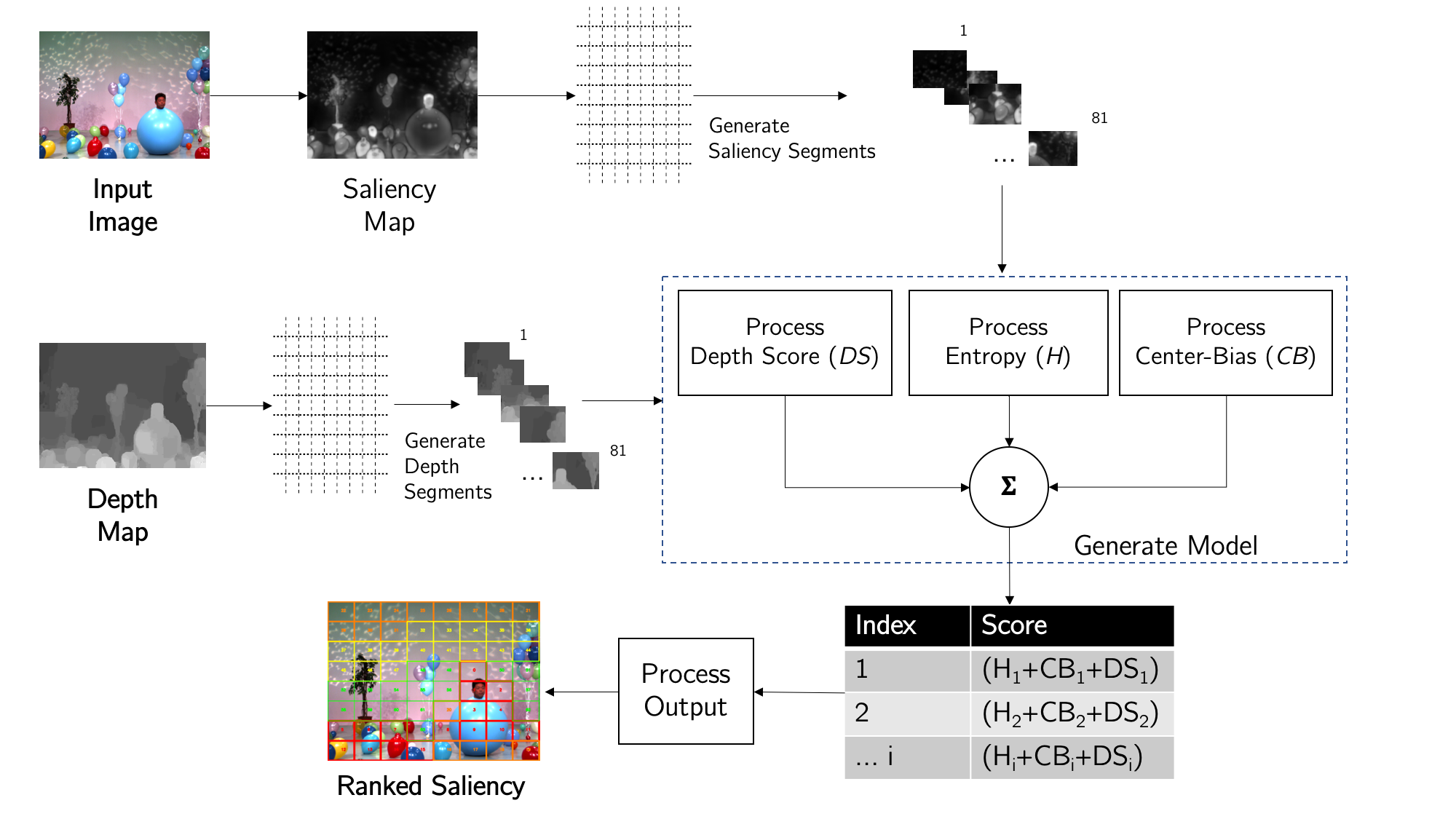}
    \caption{Saliency Ranking architecture. (Source: \cite{sara2})}
    \label{fig:sara_pipeline}
\end{figure}

Saliency Ranking is a fundamental process aimed at discerning the most visually significant features within an image. In \cite{sara, sara2}, the authors introduce a technique for separating images into segments, and ranking their saliency. Using a grid-based methodology, this technique successfully locates salient objects, that allows for automatic selection of objects without the need of training a model. SaRa operates in two separate stages. It first processes texture and, if available, depth maps with a defined grid $G$ with a size of $k \times k$, to generate a saliency map using Itti's model \cite{itti}, as implemented in \cite{pySaliencyMap}. Moreover, the presented saliency model \cite{itti}, recognises features using a constructed ``feature integration theory'' method inspired by the design and behaviour of the primate visual system. Subsequently, each segment inside grid $G$ is scored separately, taking into consideration criteria such as proximity to the image centre, accounting for centre-bias ($CB$), and entropy ($H$) computed using Equation \ref{eq:entropy} together with a depth score ($DS$). Figure \ref{fig:sara_pipeline} illustrates the aforementioned architecture.

\begin{equation}
H(X)=-\sum_{i=1}^{|t|} P\left(x_i\right) \log _2\left(P\left(x_i\right)\right)
\label{eq:entropy}
\end{equation}

The grid $G$ consisting of \(n = k^2\) segments, each yielding a corresponding saliency ranking score $S$ is calculated as depicted in Equation \ref{eqn:scoreNth}. These scores are then used to rank the segments of the input image, whereby a greater value of $S$ indicates a more salient segment in image. According to extensive testing and experimentation described in \cite{sara}, the ideal grid size for $G$ is for $k=9$, which successfully covers the region of prominent items while minimising total segment area utilisation.

\begin{equation}
\label{eqn:scoreNth}
S_n = H_n + CB_n + DS_n
\end{equation}

\subsection{Feature and Classification Learning}
\label{sec:feature_learning}

Feature Learning, particularly critical in image classification, entails extracting vital features from images to aid in accurate object and pattern detection, most notably through the use of CNNs \cite{feature_learning}. CNNs accomplish this through convolutional layers, which extract features using kernels or filters, successfully detecting patterns in images by striding across and computing dot products to generate feature maps, while pooling layers down-sample these maps to reduce parameter size. These networks have revolutionised the area of computer vision by allowing feature learning without the need to train massive complicated neuron architectures streamlining the process. % - maybe new paragraph here
To this extent, popular CNN architectures designed for image classification include: ResNet50 \cite{resnet50}, VGG16 \cite{vgg16}, Inception \cite{inception}, Xception \cite{xception}, MobileNet \cite{mobilenet}, and EfficientNet \cite{efficientnet}, each characterised by its architectural design and feature learning methods. Earlier architectures include VGG16, which emphasises depth in networks by employing lower filter sizes \cite{vgg16}, and ResNet50, which uses residual connections to develop deeper and more expressive representations \cite{resnet50}. These architectures demonstrate the varied methodologies employed in designing CNNs for feature and classification learning. On the other hand, Inception improves feature learning performance by using parallel filters of varying sizes \cite{inception, inceptionV4}. Xception takes the Inception concept to its extreme by employing depth-wise separable convolutions to efficiently capture intricate features across varying spatial scales \cite{xception}, whereas MobileNet focuses on efficient feature learning using lightweight depth-wise separable convolutions \cite{mobilenet}. With compound scaling, EfficientNet \cite{efficientnet} balances model depth, breadth, and resolution for optimal feature learning while retaining computational efficiency. Moreover, a common utilisation approach for representation learning involves leveraging these architectures pre-trained on the ImageNet dataset \cite{imagenet1, imagenet2}.

\subsection{Reinforcement Learning}
\label{sec:reinforcement_learning}
\begin{figure}[ht]
    \centering
    \includegraphics[width=0.8\linewidth]{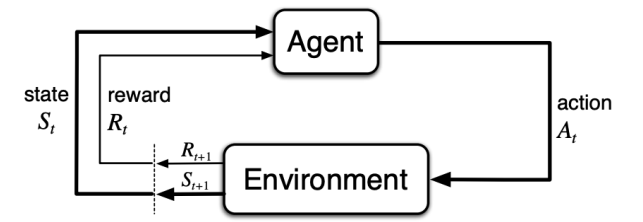}
    \caption{RL interaction loop. (Source: \cite{suttonBook})}
    \label{fig:rl_loop}
\end{figure}

Reinforcement Learning is the process of learning what to do—how to translate states into actions—in order to maximise a numerical reward signal \cite{suttonBook}. In RL, the learner is not taught which actions to perform but instead must experiment to determine which actions provide the highest return. Additionally, RL is distinguished by two key characteristics: reliance on trial-and-error search, and the capacity to maximise cumulative rewards over extended periods rather than seeking immediate satisfaction. The RL framework is often modelled as an interaction loop, as can be seen in Figure \ref{fig:rl_loop}. In the loop, the RL agent performs actions in an environment based on a policy that transitions its current state to a new one while collecting rewards and observations in the process\cite{suttonBook}.

The policy, which is frequently represented as \(\pi\), serves as a function that guides the agent's decision-making process in an environment. It maps each state \(s_t\) to an action \(A_t\), determining the agent's movements from the current state \(s_t\) to the next state $s_{t+1}$ at a given time step \(t\). \(R_t\) represents the numerical reward that the agent obtains as a result of each action in a state. These rewards are critical in refining the agent's policy, assisting it in determining the appropriate action to take in various states in order to achieve its ultimate goal: the optimal policy \(\pi^*\). Maximising the predicted cumulative reward, which is the sum of all rewards collected over multiple time steps, is required to achieve this optimum policy \cite{rl_survey}. The Environment refers to the external system or context with which an agent interacts and learns. It encompasses everything outside the agent's control, presenting states and rewarding the agent's actions, which are critical for the agent's observance, learning, and decision-making in a dynamic environment.

\subsubsection{Markov Decision Process}
\label{subsec:mdp}
Formally, RL can be expressed as a MDP \cite{deep_rl_survey}. An MDP is a mathematical framework for modelling SDMP in which the present state of the system, potential actions, and probability of transitioning to the next state are known \cite{mdp}. Moreover, an MDP can be modelled as the following tuple: $\langle S, A, P, R, \gamma \rangle$, whereby:

\begin{itemize}
    \item $S$ is a set of possible states $s_t \in S$ observed by the agent at time step $t$.
    \item $A$ is a set of possible actions $a_t \in A$ that the agent can execute at time step $t$.
    \item $P$ is the transition dynamics $p(s_{t+1}|s_t, a_t)$ that determine the probability of moving to $s_{t+1}$ after taking action $a_t$ in state $s_t$.
    \item $R$ is the immediate reward function $r(s_t, a_t, s_{t+1})$ that provides the reward obtained after transitioning from $s_t$ to $s_{t+1}$ through action $a_t$.
    \item $\gamma$ is the discount factor ($\gamma \in [0, 1]$) that regulates the balance between immediate and future rewards, and shapes the agent's decision-making.
\end{itemize}

\subsubsection{Deep Reinforcement Learning}
\label{subsec:drl}
Employing techniques such as bootstrapping, dynamic programming, monte carlo, temporal-difference, and tabular search, RL methods face the issue of handling and generalising large-scale data in broad environments. A solution to this problem pertains to the use of Deep Reinforcement Learning (DRL). DRL combines the concept of RL methods with deep learning, connecting their strengths to address shortcomings in classic tabular techniques. As described in \cite{deep_rl_survey}, DRL effectively navigates away from the confines of manually built state representations by combining complex neural network topologies such as ANNs and CNNs. It excels at exploring huge and complicated search spaces, allowing RL agents to generalise their learning across different settings. Deep RL uses Neural Networks (NN) to estimate value functions or rules, as opposed to traditional tabular techniques, which suffer from scalability in contexts with exponentially vast state spaces. This in turn allows agents to learn, and take decisions in high-dimensional and continuous state spaces, enabling them to display flexible behaviour across a wide range of complex situations \cite{deep_rl_survey}.

\subsubsection{Deep Q-Network}
\label{subsec:dqn}
Deep Q-Network (DQN) is an extension of the Q-learning technique but replaces the tabular $Q$-values with a deep neural network. In Q-learning, the $Q$-value represents the mapping that assigns a numerical value to each state-action pair, while indicates the expected cumulative reward when taking action $a$ in state $s$. Moreover, Q-learning is an off-policy temporal-difference control algorithm whereby after every time step, $Q$-values are updated using Equation \ref{eq:q_learning}.

\begin{equation}
\scalebox{0.8}{$
    Q(s, a) = Q(s, a) + \alpha \cdot \left( R + \gamma \cdot \max_{a_{t+1}} Q(s_{t+1}, a_{t+1}) - Q(s, a) \right)
$}
\label{eq:q_learning}
\end{equation}

DQN attempts to increase generalisation capabilities, particularly in cases with huge state spaces, which traditional Q-learning and memory techniques fail to control \cite{deep_rl_survey}. As illustrated in Equation \ref{eq:dqn_optimal} the DQN utilises a deep neural network in the form of a new parameter \(\theta\) which symbolises the deep neural network's weights, in order to approximate the optimal value function. To achieve this, Equation $\ref{eq:DQN_loss}$ defines the expected loss $L_i(\theta_i)$, where $U(D)$ denotes the uniform distribution over samples stored in the replay buffer $D$, ensuring a random selection of experiences for training, and $i$ represents the iteration index for the parameters $\theta_i$. Furthermore, DQN leverages this loss function to optimise the state-action value function \cite{dqn_paper}.

\begin{equation}
    Q(s, a; \theta) \approx Q^*(s, a)
    \label{eq:dqn_optimal}
\end{equation}

\begin{equation}
\scalebox{0.87}{$
\begin{aligned}
L_i(\theta_i) = & \, \mathbb{E}_{(s_t, a_t, r_t, s_{t+1}) \sim U(D)} \Bigg[ \Big( r_t + \gamma \max_{a_{t+1}} Q(s_{t+1}, a_{t+1}; \theta_i^{-}) \\
& \, \qquad \qquad \qquad \qquad \qquad \quad - Q(s_t, a_t; \theta_i) \Big)^2 \Bigg]
\end{aligned}
$}
\label{eq:DQN_loss}
\end{equation}

The deep Q-learning algorithm employed in DQN utilises both a policy network and a target network to facilitate efficient and stable learning. The policy network, or Q-network, continually approximates $Q$-values for state-action pairings in deep Q-learning to guide action selection. Similarly to normal Q-learning, the policy network updates the $Q$-values through the temporal-difference error and gradient descent. Subsequently, the target network, a separate network with fixed parameters updated from the policy network on a regular basis, stabilises learning by providing consistent $Q$-value objectives during training \cite{dqn_paper}. As previously hinted in Equation \ref{eq:DQN_loss}, DQN also employs an experience Replay Buffer $RB$ , which can be used to sample through transition data during training in order to facilitate more stable learning. These transitions consist of a tuple of five elements containing: state, action, reward, done, and after state. Transitions in the environment are added to a $RB$, and the loss and gradient are calculated using a batch of transitions sampled from the $RB$ rather than utilising the most recent transition while training \cite{dqn_paper, suttonBook}. Considering that DQN is an extension of Q-learning, it also employs the $\epsilon-$Greedy policy to select an action. Thus, ensuring a balance between agent exploration and exploitation, whereby the optimal action is chosen with a probability of $1- \epsilon$ otherwise, a random action is selected \cite{suttonBook, deep_rl_survey}.

\subsubsection{Double Deep Q-Network}
\label{subsec:ddqn}
The Double Deep Q-Network (DDQN) algorithm is an improvement on the traditional DQN approach, designed to address overestimation difficulties in Q-learning methods and reducing maximisation bias \cite{double_dqn_paper}. DDQN works by separating action selection from action value estimation using two independent NN. In DDQN, the network with the highest $Q$-value determines the ideal action, whereas the alternate network evaluates the value of that action. The difference is apparent when comparing the target value functions in Equations \ref{eq:dqn_compare} and \ref{eq:double_dqn_compare}. In these equations, $Y_t$ represents the target value at time step $t$, used in the update rule, and composed of the immediate reward $R_{t+1}$ and a discounted $Q$-value. In Equation \ref{eq:dqn_compare}, the $Q$-value is maximised using the same network parameters $\theta_t$ for both action selection and value estimation. Conversely, in Equation \ref{eq:double_dqn_compare}, a separate network with target network parameters $\theta_t^{-}$ is utilised for value estimation. Significantly, this improves the DDQN algorithm's stability and learning efficiency by limiting the overestimation bias found in single-network Q-learning techniques \cite{deep_rl_survey}.  

\begin{equation}
\scalebox{0.8}{$
Y_t^{\mathrm{DQN}} = R_{t+1} + \gamma Q\left(s_{t+1}, \underset{a}{\operatorname{argmax}} Q\left(s_{t+1}, a ; \theta_t\right) ; \theta_t\right)
$}
\label{eq:dqn_compare}
\end{equation}
\begin{equation}
\scalebox{0.8}{$
Y_t^{\mathrm{DoubleDQN}} \equiv R_{t+1} + \gamma Q\left(s_{t+1}, \underset{a}{\operatorname{argmax}} Q\left(s_{t+1}, a ; \theta_t\right), \theta_t^{-}\right)
$}
\label{eq:double_dqn_compare}
\end{equation}

\subsubsection{Dueling Deep Q-Network}
\label{subsec:duelingdqn}
The Dueling Deep Q-Network (Dueling DQN) architecture presents a substantial improvement on the traditional DQN. It improves efficient learning whilst allowing the model a more precise identification of the crucial state values, and advantageous actions \cite{deep_rl_survey}. Moreover, Dueling DQN presents a revolutionary design that separates the estimation of state values from the assessment of each action taken, independently of the remaining state value in a specific state \cite{dueling_dqn_paper}.

\subsubsection{Double Dueling Deep Q-Network}
\label{subsec:doubleduelingdqn}
The Double Dueling Deep Q-Network (D3QN) architecture represents a significant advancement over both traditional DQN and Dueling DQN models, by enhancing learning efficiency and decision-making accuracy in DRL tasks \cite{rl_survey, suttonBook}. Through the integration of DDQN and Dueling DQN, D3QN presents a dual-network architecture that separates the estimation of state values, and action advantages. Additionally, double $Q$-learning techniques are utilised to reduce overestimation biases \cite{rl_survey, suttonBook}. This novel technique not only increases the model's capacity to discover critical state-action values, but also makes it much more resilient to sub-optimal action selection, rendering it ideal for complex and dynamic contexts.
\section{Literature Review}%
\label{chp:lit_review}
This chapter will focus on the literature review necessary for understanding the subsequent chapters. It will identify key architectures and methodologies previously developed in reinforcement learning object detectors.

Pioneering reinforcement learning-based object detectors, and appearing around the same time RCNN was released (2015), Caicedo et al. present ``Active Object Localization with Deep Reinforcement Learning''. In their study \cite{object_localization_RL}, DRL was utilised for active object localisation, whereby the object detection problem was transformed into the MDP framework, in an attempt to solve it. In their implementation, the authors took into consideration eight distinct actions (up, down, left, right, bigger, smaller, fatter, and taller) to enhance the bounding box's fit around the item and an extra action (trigger) to indicate that an object is correctly localised in the current box. The trigger also modifies the environment by using a black cross to mark the region covered by the box, which acts as an Inhibition of Return (IoR) mechanism to prevent re-attention to the currently attended region, a widely used strategy in visual attention models to suppress continuous attraction to highly salient stimuli \cite{itti_ior}. Furthermore, in their approach, the authors tackled multiple object detection by allowing the RL agent to run for a maximum of 200 time steps, evaluating 200 regions for each given image. Caicedo et al. encoded the state as a tuple of feature vectors and action histories, while also employing shift in Intersection over Union (IoU) as a binary reward strucuture ($[-1, +1]$) across actions, and a subsequent reward for trigger actions. The authors also employed a guided exploration strategy based on apprenticeship learning principles \cite{apprentice_learning_1, apprentice_learning_2, apprentice_learning_3}, which uses expert demonstrations to inform the agent's actions, particularly in determining positive and negative rewards based on IoU with the current bounding box, allowing the agent to choose positive actions at random during exploration.

An enhancement to the algorithm presented in \cite{object_localization_RL} was put forth by \cite{hierarchicaldetection}, whose authors also treated the object detection problem as an MDP and employed a hierarchical technique for object detection. The RL agent's objective in their approach consisted of first identifying a Region of Interest (ROI) in the image, then slowly reducing the area of interest to obtain smaller regions, thus generating a hierarchy. Similar to \cite{object_localization_RL}, Bellver et al. \cite{hierarchicaldetection} additionally represented the reward function as an IoU change in between actions. According to their paper, they utilised DQN as the agent's architecture while also employing Image-zooms and Pool45-crops with VGG16 \cite{vgg16} networks as the image feature extraction backbone. In addition to a memory vector that stores the last four actions, the extracted image feature vector from the aforementioned networks was incorporated into the DQN state.

\begin{figure}[ht]
    \centering
    \includegraphics[width=0.9\linewidth]{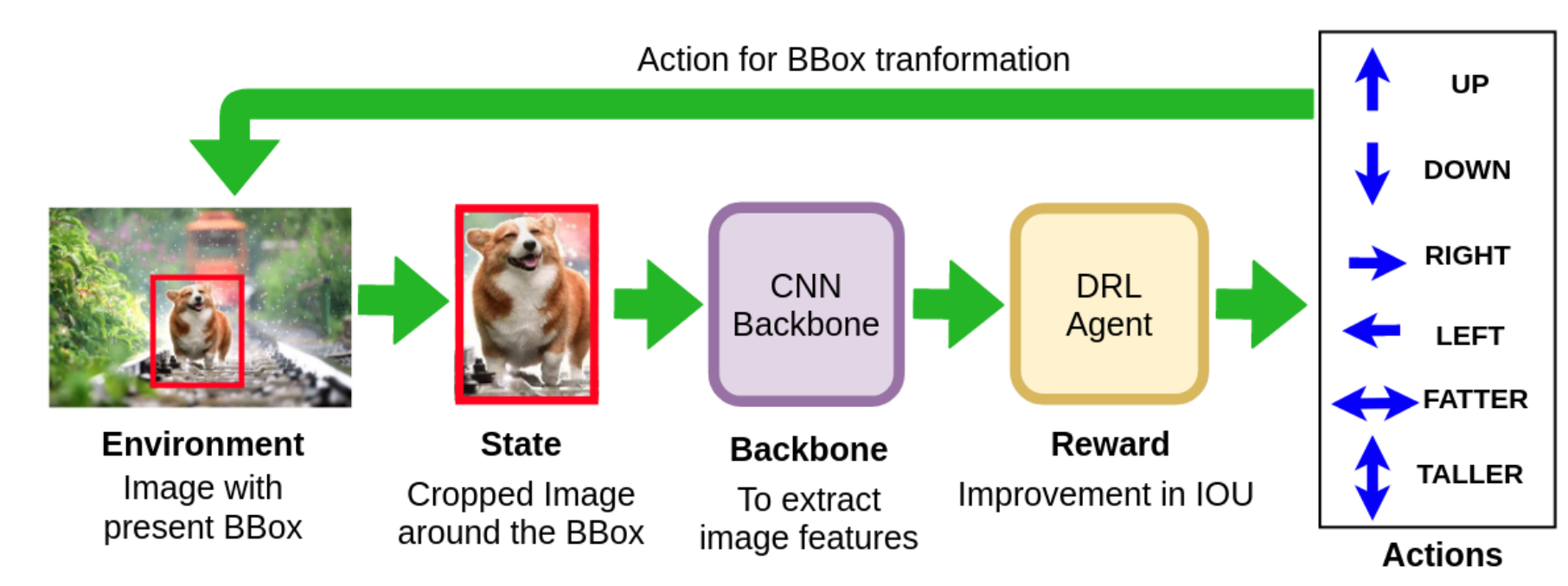}
    \caption{DRL implementation for object detection. (Source: \cite{deepRLinCV})}
    \label{fig:basic_drl_object_detection_pipeline}
\end{figure}

K. Simonyan et al. \cite{detectionSequential}, use the sequential search technique to present a DRL-based object identification system. The model was trained by using a collection of image regions, where the agent returned fixate actions at each time step, identifying an area in the image that the actor should investigate next. The state was composed of three pieces in total: the focus history $F_t$, the selected evidence region history $E_t$, and the observed region history $H_t$. The fixate action was composed of three elements: the image coordinate of the subsequent fixate $z_t$, the index of the evidence region $e_t$, and the $fixate$ action itself. As presented in their paper, K. Simonyan et al. specified the reward function to be responsive to the detection location, the confidence of the final state, and the imposition of a penalty for each region evaluation.

As described in  \cite{treeStructuredRL}, Z. Jie et al. proposed a tree-structured RL agent (Tree-RL) for object localisation in order to map the inter-dependencies among the various items. By treating the problem as an MDP, the authors took into account two different kinds of actions: translation and scaling. The former consisted of five actions, while the latter involved eight. In addition, the agent's defined state consists of a concatenation of the extracted feature vectors from the current window, the entire image, and the history of actions taken. The feature vector was constructed from a pre-trained VGG16 \cite{vgg16} model on the ImageNet \cite{imagenet1} \cite{imagenet2} dataset, whilst also incorporating IoU shifts between actions as a reward function. The proposed Tree-RL agent architecture uses a top-down tree search, which starts with the entire image and proceeds to iteratively select the optimal action from each action group, thereby generating two new windows. This process is performed recursively in order to successfully locate the object in the image.

Contrary to previous techniques, \cite{multitaskLearning} suggested a multitask learning policy that uses DRL for object localisation. Y. Wang et al. also viewed the problem as a MDP, wherein the agent needed to execute a series of actions to manipulate the bounding box in various ways. A total of eight actions for bounding box transformation (left, right, up, down, bigger, smaller, fatter, and taller) were employed.  Similar to the previously mentioned methods, the authors encoded the states as a fusion of historical actions and feature vectors while also employing shift in the IoU as a reward between actions. An increase in IoU would result in a reward of 0, while a decrease to this value would result in a reward of -1. On the other hand, the reward for the terminal action was set to 8 if the IoU was larger than 0.5 and -8 otherwise. Moreover, in their paper, Y. Wang et al. separated the eight transformation actions and the terminal action into two networks and trained them together using DQN with multitask learning for object localisation.
Incorporating RL, A. Pirinen and C. Sminchisescu in \cite{rpn_2018_CVPR} suggested an improvement to Regional Proposal Networks (RPN) that greedily chooses ROIs. In this research, the authors utilised a two-stage detector comparable to Fast RCNN \cite{fastrcnn} and Faster RCNN \cite{fasterrcnn}, whilst also employing RL in the decision-making process. They also utilised the normalised shift in IoU as the reward.

Rather than learning a policy from a huge quantity of data, \cite{barRL} presented a bounding box refinement (BAR) technique based on RL. In their publication, M. Ayle et al. utilised the BAR algorithm to anticipate a sequence of operations for bounding box refinement for imprecise bounding boxes predicted by some algorithm. Similarly to previous methods, they examined eight different bounding box transformation actions (up, down, left, right, broader, taller, fatter, and thinner) and categorised the problem as a SDMP. They suggested two methods: BAR-DRL, which is an offline algorithm, and BAR-CB, which is an online algorithm and requires training on every image. The developers of BAR-DRL trained a DQN over the states using a history vector of ten actions and features taken from ResNet50 \cite{resnet50} \cite{inceptionV4}, which were pre-trained on ImageNet \cite{imagenet1} \cite{imagenet2}. The encoded reward for the BAR-DRL agent after executing an action consisted of 1 in the case of an IoU increase and -3 otherwise. To record the shape and boundaries of the object of interest, M. Ayle et al. modified the LinUCB \cite{linucb} method for BAR-CB and took into consideration the Histogram of Oriented Gradients (HOG) for the state. Additionally, the online technique (BAR-CB) followed the same steps as the offline method, with a 1 reward for a positive shift in IoU, and a 0 reward otherwise. Moreover, the authors regarded $\beta$ as the terminal IoU for both implementations.

A structure including two modules, coarse and fine-level search, was proposed by \cite{uzkent2020efficient} as an enhancement to the sequential search approach of \cite{detectionSequential}. In their study, B. Uzken et al. claim that their approach is effective for detecting objects in huge images with dimensions exceeding 3000 pixels. The authors initiated their process with a broad-scale examination of a large image, identifying a set of patches. These patches were subsequently employed in a more detailed analysis to locate sub-patches. A two-step episodic MDP was used for both fine and coarse levels, with the policy network's job being to return the probability distribution of each action. Additionally, the authors represented the action space using a binary array format ($[0, 1]$), where a value of 1 denotes the agent's decision to explore obtaining sub-patches for a specific patch, and 0 indicates otherwise. As described in their paper, B. Uzken et al. applied the linear combination of $R_{acc}$ (detection recall) and $R_{cost}$ (image acquisition cost plus run-time performance reward) in their design, accounting for a total of 16 patches and 4 sub-patches, respectively.

M. Zhou et al. presented ReinforceNet \cite{ReinforceNet}, a DRL framework for object detection. Formulating the object detection challenge as an MDP, the authors suggested an agent that iteratively alters an initial bounding box by executing actions such as translate, scale, and aspect ratio change. In addition, actions are parameterised using a deep policy network based on the VGG16 \cite{vgg16} feature extractor network. The encoded state was made up of features from ROI-pooling CNN layers as well as a history vector. The reward was designed as a change in IoU after each action, and the episode ends when the IoU exceeds a certain threshold. The REINFORCE algorithm was used to train an end-to-end policy network using a learned value function baseline. The authors also introduced a ``completeness'' metric in their paper to evaluate the extent to which the agent explores the state space during training. Low completeness shows that the agent converges prematurely without sufficient exploration. M. Zhou et al. identified inadequate exploration as a major difficulty in applying RL for object localisation, as well as implementing a curriculum learning technique with the aim of improving completeness and exploration.

M. Samiei and R. Li \cite{samiei2022object} propose two DRL methods to actively solve the object detection problem. Similarly to \cite{hierarchicaldetection}, the authors proposed a hierarchical method that enables an RL agent to select in a timely manner, one out of five sub-regions of the image, with the aim of obtaining smaller regions. Additionally, the proposed dynamic method allows the agent to transform the bounding box at each time step by using actions like translate, scale, and deform. M. Samiei and R. Li also formalised the problem as an MDP and encoded the state as a combination of the current region features and action history. In their study, the authors also experimented with different reward functions, examining both IoU improvements and recall as measurements for this function. In a comparable manner to \cite{hierarchicaldetection}, the authors estimated $Q$-values using the DQN architecture with an experience RB. As an image feature extraction backbone, the VGG16 \cite{vgg16} was also used. Unfortunately, the proposed method has limitations such as class-specific training and struggles with multiple objects.

Recent advancements on the approach discussed in \cite{object_localization_RL}, however tackling single object detection, encompass innovations highlighted in \cite{dissertation, rapport}. In the study presented in \cite{dissertation}, a novel approach utilising Q-learning with CNNs is introduced to devise a strategy for object detection via visual attention. The study showcases innovative experiments that dismiss pre-trained networks for representation learning. In addition, it demonstrates the network's capability to generalise within and across super-categories, detect objects of any type, and achieve comparable results to previous research. This master's thesis investigates the inclusion of two split-view actions and evaluates an incremental training technique. Additionally, it tests whether fine-tuning the model for one category and then bootstrapping training for the next category is better than training on a shuffled dataset involving all categories. On the other hand, in the implementation discussed in \cite{rapport}, the author addresses the challenge of single object detection and explores the utilisation of the VGG16 network \cite{vgg16} for representation learning, deviating from the 6-layer CNN specified in the original paper \cite{object_localization_RL}, yet achieving somewhat comparable outcomes.

\section{Methodology}%
\label{chp:methodology}

To identify the most relevant object inside an image, this system combines a saliency ranking algorithm with reinforcement learning. It seeks to detect and identify critical aspects in visual data by integrating various methodologies. This chapter describes the implementation details of the proposed "SaRLVision" system architecture, explaining the design decisions and their justifications. The approach and techniques used in creating the system architecture are also described, drawing on knowledge gained from thorough research of pertinent literature. Moreover, all code utilized is made available on GitHub\footnote{\url{https://github.com/mbar0075/SaRLVision}}.

\subsection{Reinforcement Learning}
\label{sec:rl_methodology}
In alignment with the primary goal, an RL framework was developed to achieve object localisation within images. To this extent, the developed system was built via the \texttt{gymnasium}\footnote{\url{https://gymnasium.farama.org/index.html}} API, which provided a platform for the problem formulation, inspired by the existing literature. Subsequently, DRL techniques were applied to approximate the object detection problem.

\subsubsection{States}
\label{subsec:states}

Similar to methodologies presented in several prior studies \cite{object_localization_RL, multitaskLearning, dissertation, rapport}, the state representation in this work comprises a tuple $\langle o, h \rangle$, where $o$ denotes a feature vector corresponding to the observed region (current bounding box region), and $h$ encapsulates the history of actions undertaken. 
Departing from the methodologies delineated in \cite{object_localization_RL, dissertation}, the evaluation of the developed system involved testing three distinct image feature extraction methods (initialised with the ImageNet weights): VGG16, ResNet50, and MobileNet, facilitated by the \verb|pytorch|\footnote{\url{https://pytorch.org/}} API. VGG16 and ResNet50 were selected due to their widespread adoption in the literature, while MobileNet was chosen for its lightweight architecture. However, unlike the approaches described in \cite{rapport, multitaskLearning}, the feature maps undergo processing through a global average pooling layer, a technique utilised in both Inception \cite{inceptionV4}, and Xception \cite{xception} architectures. This step serves to reduce dimensionality and expedite object detection, while facilitating the development of smaller models and preserving the essential features from the image without compromising quality through Dense layers.
Notably, pre-trained models were favoured over training a convolutional DQNfrom scratch, as demonstrated in \cite{dissertation}. These pre-trained models had already been trained on larger datasets, and interestingly, \cite{dissertation} showed that utilising a pre-trained CNN achieved comparable results.
Additionally, akin to the method introduced in \cite{object_localization_RL}, the history vector $h$ is a binary vector encoding the ten most recent actions taken, implemented to mitigate the repetition of similar or opposing actions. Each action in the history vector is represented by a nine-dimensional binary vector. A comparative analysis of different state sizes is presented in Table \ref{tab:state_size}.

% So dqn state size: \cite{object_localization_RL} = 4096 + 90 = 4186 (90 action history)
% \cite{rapport} = 25088 + 90 = 25178
% VGG16 = 512 + 90 = 602
% ResNet50 = 2048 + 90 = 2138
% MobileNet = 1280 + 90 = 1370
% So for the feature extraction networks I am extracting features then applying a global average pooling (InceptionV3, Xception both use it)

\begin{table*}[ht]
    \centering
    % \footnotesize
    \begin{tabular}{|c|c|c|c|c|c|}\hline
         \textbf{Architecture}&  \textbf{VGG16 (Ours)}&  \textbf{MobileNet (Ours)}&  \textbf{ResNet50 (Ours)}&  \textbf{CNN (\cite{object_localization_RL})}& \textbf{VGG16 (\cite{rapport})}\\ \hline \hline
         \textbf{State Size}&  602&  1370&  2138&  4186& 25178\\ \hline
    \end{tabular}
    \caption{Comparison of state sizes.}
    \label{tab:state_size}
\end{table*}% Global average pooling plus action history size

\subsubsection{Actions}
\label{subsec:actions}

\begin{figure}[ht]
    \centering
    \includegraphics[width=0.85\linewidth]{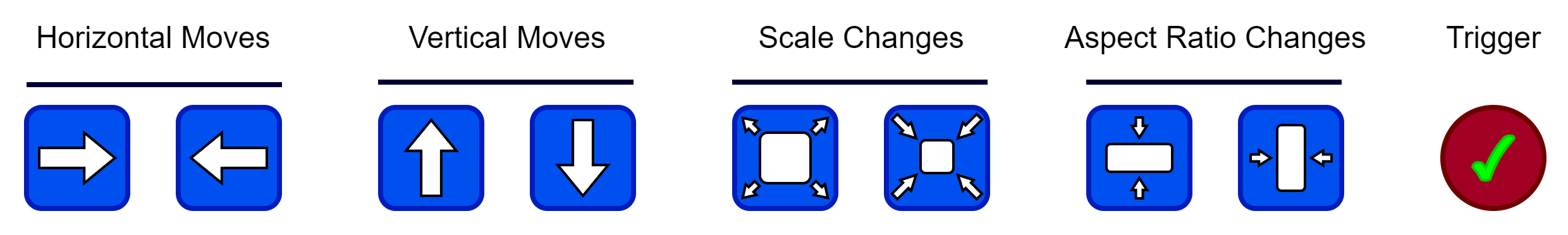}
    \caption{Set of possible actions, inspired from \cite{object_localization_RL}.}
    \label{fig:actions}
\end{figure}

In a manner akin to the methodologies described in \cite{object_localization_RL, rapport, multitaskLearning}, the action set \( A \) comprises of eight transformations applicable to the bounding box, alongside another action designed to terminate the search process. These transformations, illustrated in Figure \ref{fig:actions}, are categorised into four subsets: horizontal and vertical box movements, scale adjustment, and aspect ratio modification. Consequently, the agent possesses four degrees of freedom to adjust the bounding box coordinates ([\( x_1, y_1, x_2, y_2 \)]) during interactions with the environment. 

% N.B. to get this good diagram had to do export zoom in draw.io to 400%

Each transformation action induces a discrete change to the box's size relative to its current dimensions, influenced by the aspect ratio, as expressed in Equation \ref{eqn:aspect_ratio}.
\begin{equation}
\label{eqn:aspect_ratio}
\alpha_w=\alpha *\left(x_2-x_1\right) \quad \alpha_h=\alpha *\left(y_2-y_1\right)
\end{equation}
While \cite{dissertation} introduces two additional split view actions, these were omitted, to streamline complexity, as their transformations could be replicated using existing actions.
In alignment with the methodologies outlined in \cite{object_localization_RL, dissertation, rapport}, the parameter $\alpha$ is established at 0.2, as a smaller value resulted in slower transformations. Moreover, a trigger action is incorporated to indicate successful object localisation by the current box, thereby concluding the ongoing search sequence, and drawing an IoR marker on the detected object \cite{itti_ior}, similar to \cite{object_localization_RL}. This approach mirrors the strategies employed in \cite{dissertation, rapport}, particularly in single object detection scenarios.

\subsubsection{Rewards}
\label{subsec:rewards}

The reward function (\( R \)) quantifies the agent's progress in localising an object following a specific action, assessing improvement based on the IoU between the target object and the predicted box. This function, as described in \cite{object_localization_RL, dissertation, multitaskLearning, rapport}, is computed during the training phase using ground truth boxes. For an observable region with box \( b \) and ground truth box \( g \), the IoU is calculated as shown in Equation \ref{eqn:iou}. 
\begin{equation}
\label{eqn:iou}
\text{IoU}(b, g) = \frac{{\text{area}(b \cap g)}}{{\text{area}(b \cup g)}}
\end{equation}
Equation \ref{eqn:normal_reward} defines the reward \( R_a(s_t, s_{t + 1}) \), that is assigned to the agent when transitioning from state \( s_t \) to \( s_{t + 1} \), where each state \( s \) corresponds to a box \( b \) containing the attended region, and \( t \) denotes the current time step.
\begin{equation}
\label{eqn:normal_reward}
R_a\left(s_{t}, s_{t + 1}\right)=\operatorname{sign}\left(\operatorname{IoU}\left(b_{t + 1}, g\right)-\operatorname{IoU}(b_t, g)\right)
\end{equation}
The reward is determined by the sign of the change in IoU, encouraging positive rewards for improvements and negative rewards for reductions in IoU. This binary reward scheme \( R_a(s_t, s_{t + 1}) \in \{-1, +1\} \) applies to any action transforming the box, ensuring clarity in the agent's learning process. Moreover, to minimise redundant actions, in case the reward is 0 indicating no change, a negative reward is issued.
The trigger action employs a distinct reward scheme due to its role in transitioning to a terminal state without altering the box, resulting in a zero differential IoU. Its reward is determined by a threshold function of IoU, expressed in Equation \ref{eqn:trigger_reward}, where $\eta$ represents the trigger reward and is set to 3.0 based on experiments in \cite{object_localization_RL}. The parameter $\tau$ denotes the minimum IoU required to classify the attended region as a positive detection, typically set to 0.5 for evaluation, but adjusted to 0.6 during training, in alignment with \cite{object_localization_RL}, to encourage better localisation. An enhancement to the pipeline in \cite{object_localization_RL, dissertation, rapport}, involves multiplying the trigger reward by \(2 \times \text{current IoU}\) to ensure a minimum reward of 3.0 ($\eta$) while allowing for variable rewards based on the final IoU, thereby providing the RL agent with additional incentive to improve prediction accuracy.
\begin{equation}
\label{eqn:trigger_reward}
R_\omega\left(s_{t}, s_{t + 1}\right)= \begin{cases}+\eta * 2 * \operatorname{IoU}(b, g) & \text { if } \operatorname{IoU}(b, g) \geq \tau \\ -\eta & \text { otherwise }\end{cases}
\end{equation}

\subsubsection{Episodes}
\label{subsec:episodes}

In terms of episode structure, each episode is restricted to a maximum of 40 time steps, in alignment with previous research \cite{dissertation, rapport, object_localization_RL}. Additionally, applying the trigger action prematurely also results in terminating the episode. Upon completion of an episode for a particular image, the DRL agent provides a bounding box prediction for the detected object.

\subsubsection{Network Architecture}
\label{subsec:architecture_methodology}
\begin{figure}[ht]
    \centering
    \includegraphics[width=0.85\linewidth]{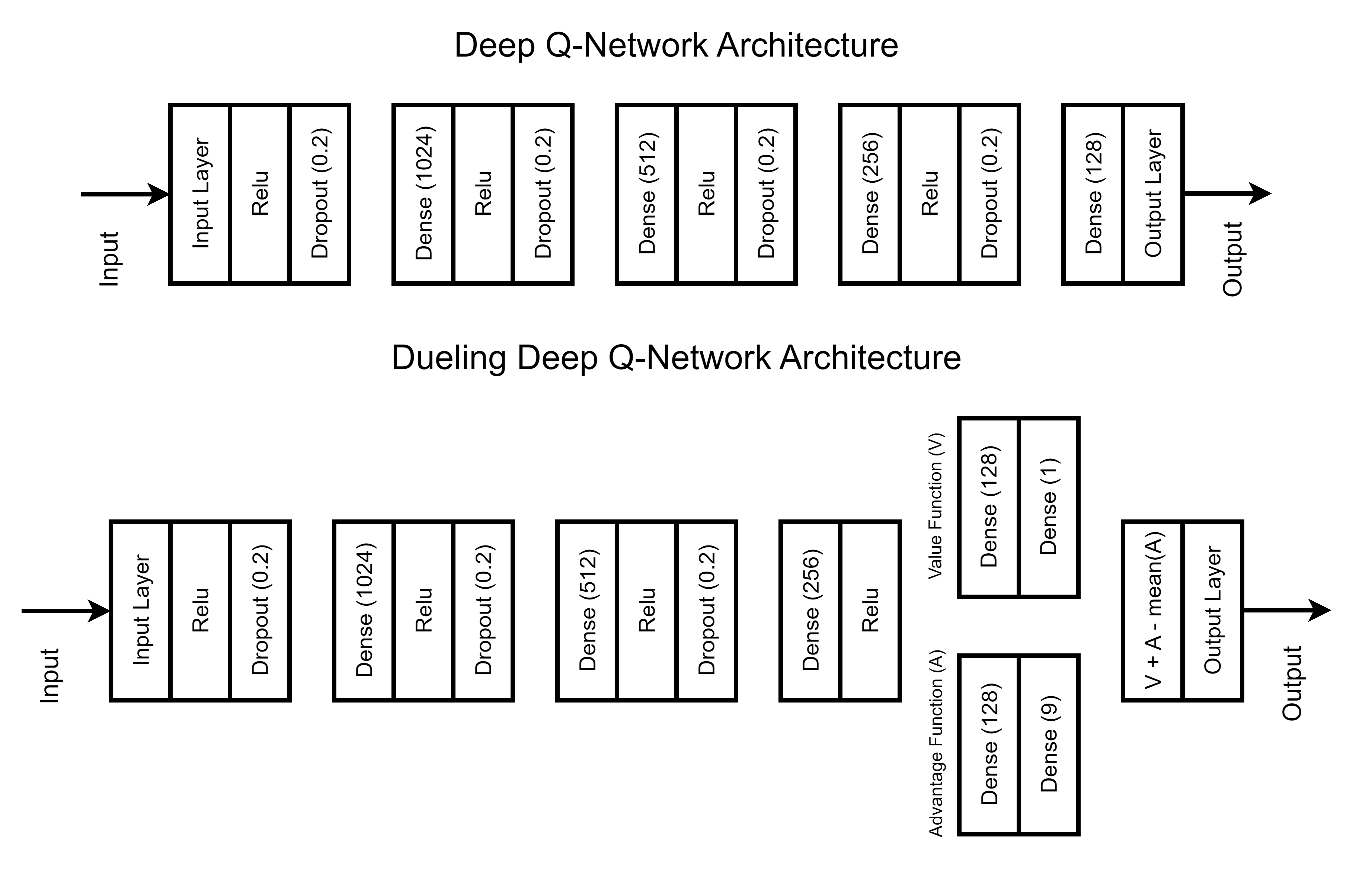}
    \caption{Proposed Deep Q-Network and Dueling Deep Q-Network architectures.}
    \label{fig:dqn_architectures}
\end{figure}

The DQN architecture, introduced in the presented system, assumes responsibility for decision-making in object localisation. To this extent, the designed architecture draws inspiration from the methodologies outlined in \cite{object_localization_RL, rapport}. While \cite{object_localization_RL} employs a 3-layer Dense Layer DQN representation, \cite{rapport} extends this architecture by incorporating Dropout Layers. Our proposed approach, as illustrated in Figure \ref{fig:dqn_architectures} which was implemented using the \verb|pytorch| API, advocates for a deeper DQN network to bolster decision-making capabilities and enhance learning complexity. To mitigate concerns regarding overfitting, Dropout Layers were integrated into the network architecture, following insights gained from the results presented in \cite{dissertation}.
Furthermore, as shown in Figure \ref{fig:dqn_architectures}, this work develops a Dueling DQN Agent to improve learning efficiency by decoupling state and advantage functions. The Dueling DQN design divides the $Q$-value function into two streams, allowing the agent to better comprehend the value of doing specific actions in different situations, as shown in \cite{dueling_dqn_paper}.
To achieve better results, the proposed approach not only introduces Dueling DQN, a technique not explored in existing literature, but also implements DDQN and D3QN methods, both of which have not been previously examined.
Additionally, while \cite{object_localization_RL} adopts a strategy of employing class-specific DQN agents, \cite{dissertation} investigates retraining class-specific DQNs on additional categories. However, this research focused solely on single object detection using class-specific DQN agents and refrained from following the experimentation listed in \cite{dissertation}.

% N.B. to get this good diagram had to do export zoom in draw.io to 400%

\subsubsection{Parameters}
\label{subsubsec:params}

Similar to the parameter choices in \cite{object_localization_RL}, the following settings were adopted for training the DRL agents in this study. The learning rate, denoted by $\alpha_l$, was set to $1 \times 10^{-4}$, while the discount factor, denoted by $\gamma$, was adjusted to $0.9$. Furthermore, the $\epsilon$-Greedy exploration strategy was implemented with initial exploration probability ($\epsilon_{\text{start}}$) of $1.0$, final exploration probability ($\epsilon_{\text{end}}$) of $0.01$, and an epsilon decay rate ($\epsilon_{\text{decay}}$) of $0.999$. These parameter values were chosen to balance exploration and exploitation during training, similar to previous works in the field \cite{object_localization_RL, rapport, dissertation, multitaskLearning}.

\subsubsection{Exploration}
\label{subsubsec:exploration}

The study explores two distinct exploration strategies for evaluation: random exploration and guided exploration. In random exploration, actions are selected using the $\epsilon$-Greedy policy, which balances between agent exploration and exploitation. This policy ensures that the optimal action is chosen with a probability of $P (1 - \epsilon)$, while otherwise a random action is selected. Guided exploration, inspired by principles from apprenticeship learning \cite{apprentice_learning_1, apprentice_learning_2, apprentice_learning_3} as outlined in \cite{object_localization_RL}, involves leveraging demonstrations made by an expert to the agent. Since the environment possesses ground truth boxes and the reward function is calculated based on the IoU with the current bounding box, positive and negative rewards for actions can be identified. Utilising this knowledge, the agent is allowed to select at random from the list of positive actions during exploration. If this set is empty, the agent may select any action. 

\subsection{Saliency Ranking}
\label{sec:sara_methodology}
\begin{figure}[!htbp]
    \centering
    \includegraphics[width=0.8\linewidth]{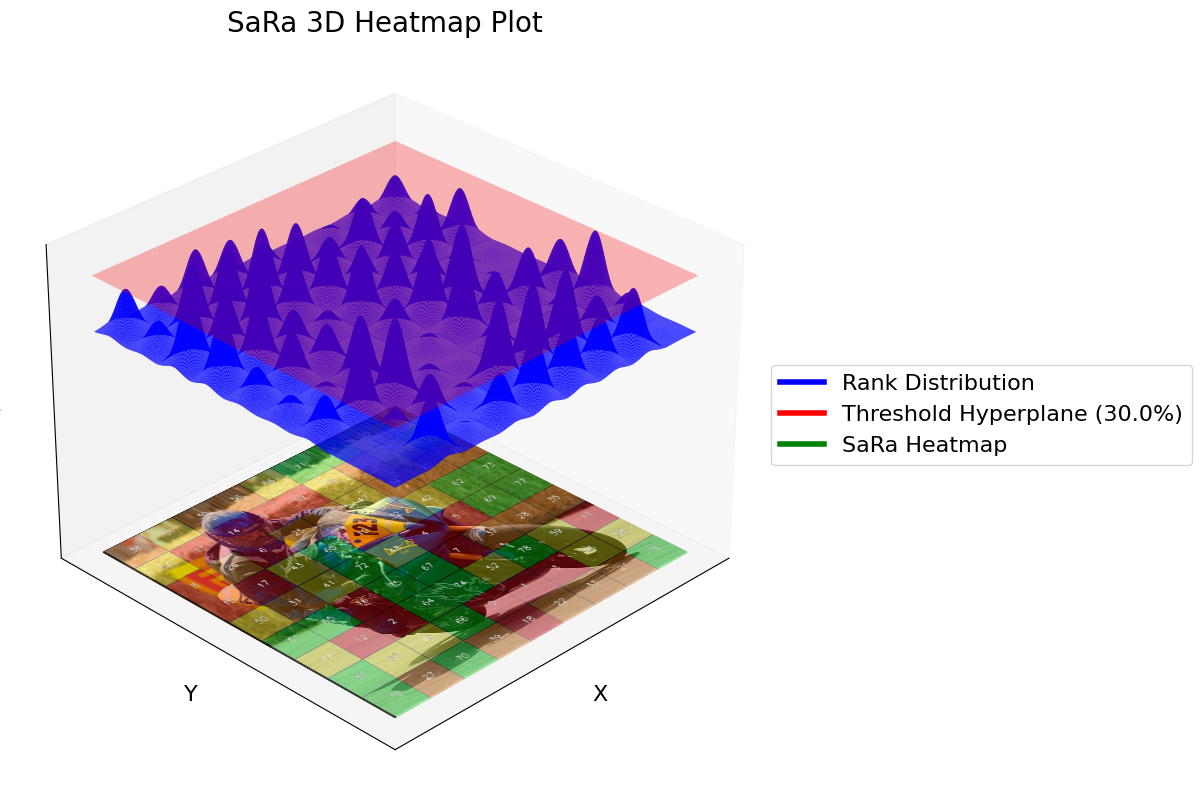}
    \caption{3D visual illustration for extracting most important ranks from saliency ranking.}
    \label{fig:ranking3D}
\end{figure}

Through the integration of saliency ranking into the developed system, an initial bounding box prediction could be achieved. Alternatively, users may choose not to employ this technique, resulting in the initial bounding box covering the entirety of the input image, a practice commonly observed in existing literature \cite{rapport, dissertation, object_localization_RL, treeStructuredRL, multitaskLearning}.
Following the acquisition of the Saliency Ranking heatmap from SaRa \cite{sara}, the first stage of this process entails the extraction of a bounding box that delineates the pertinent image segments. This technique considers a proportion of the highest-ranked areas, with a fixed threshold of 30\% and number of iterations set to 1, selected after significant experimentation (see Section \ref{sec:experiment1}). The generation of these initial bounding boxes is critical due to the fact that it allows for the separation and delineation of prominent regions in the image, for further refinement utilising RL techniques.
A visual illustration of this technique is presented in Figure \ref{fig:ranking3D}, whereby a linear hyperplane threshold is used to select the important ranks preserved for the construction of the initial bounding box prediction.

\subsection{Object Classification}
\label{sec:classification_methodology}

The inclusion of object classification significantly enhances the object detection process by providing both a label and a confidence score for identified objects. This feature continues to supply crucial contextual details about the detected object. Notably, earlier works such as \cite{object_localization_RL, rapport, dissertation, multitaskLearning, treeStructuredRL} lacked this capability, whereas the developed system provides such functionality. Moreover, the system offers versatility in the selection of architectures, allowing users to utilise a diverse array of options including ResNet50 \cite{resnet50}, VGG16 \cite{vgg16}, Inception \cite{inception}, Xception \cite{xception}, MobileNet \cite{mobilenet}, and EfficientNet \cite{efficientnet} through the \verb|keras|\footnote{\url{https://keras.io/}} API, thereby providing comprehensive options for object detection tasks.

\subsection{Transparency}
\label{sec:visual_methodology}

\begin{figure}[!htbp]
    \centering
    \includegraphics[width=\linewidth]{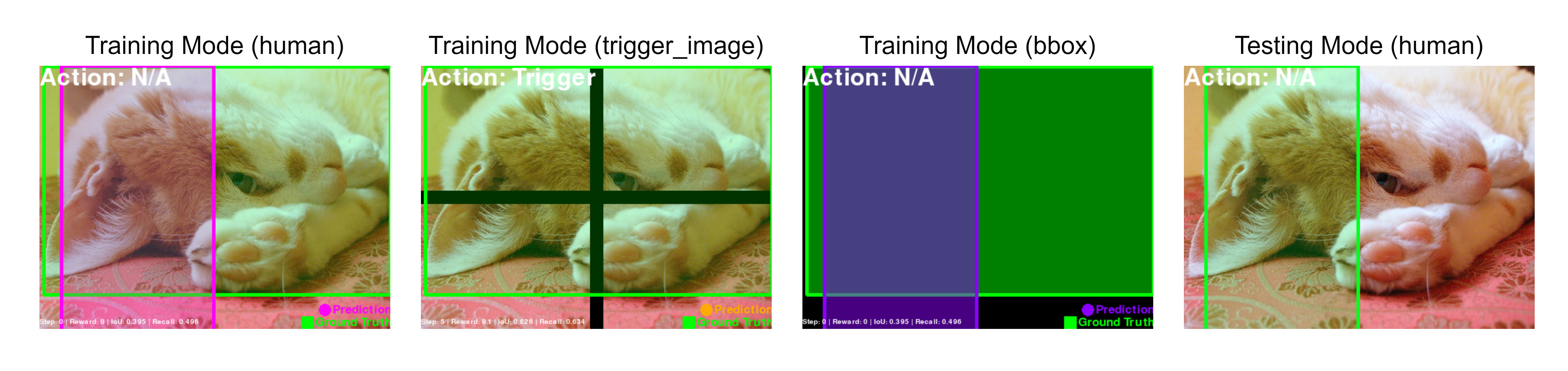}
    \caption{Different render mode visualisations.}
    \label{fig:visualisations}
\end{figure}

The study also proposes a system that creates a log and displays the current environment in several rendering modes to illustrate transparency, as demonstrated in Figure \ref{fig:visualisations}. 

% N.B. to get this good diagram had to do export zoom in draw.io to 400%

\begin{figure}[!htbp]
    \centering
    \includegraphics[width=\linewidth]{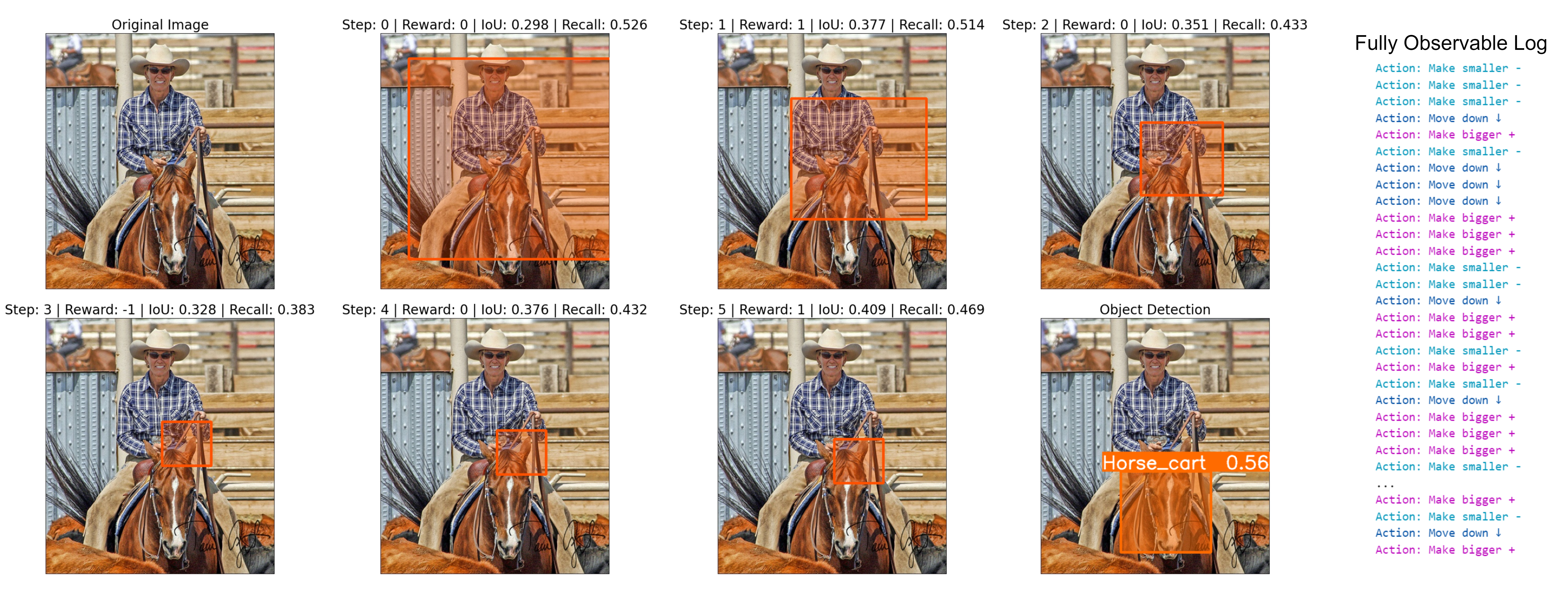}
    \caption{Functionality of the fully observable action log.}
    \label{fig:explainability}
\end{figure}

% \noindent
These visualisations provide users with insights into the current action being performed, the current IoU, the current Recall as described in \cite{hierarchicaldetection}, the environment step counter, the current reward, and a clear view of the current bounding box together with the ground truth bounding box locations in the original image. Additionally, these visualisations were designed to be dynamic across various image sizes. Unlike all object detectors and methodologies previously discussed, this methodology permits decision-making observation during the training phase, subject to a slight time overhead (a few seconds for each image) for the creation of visualisations through the \verb|gymnasium| and \verb|pygame|\footnote{\url{https://www.pygame.org/news}} APIs. 
Nonetheless, the system provides a clear action log outlining the framework's decision-making process for the current object being detected. This allows insight into the object detector's training and assessment, as depicted in Figure \ref{fig:explainability}.

\subsection{System Overview}
\label{sec:system overview}

\begin{figure}[!htbp]
    \centering
    \includegraphics[width=\linewidth]{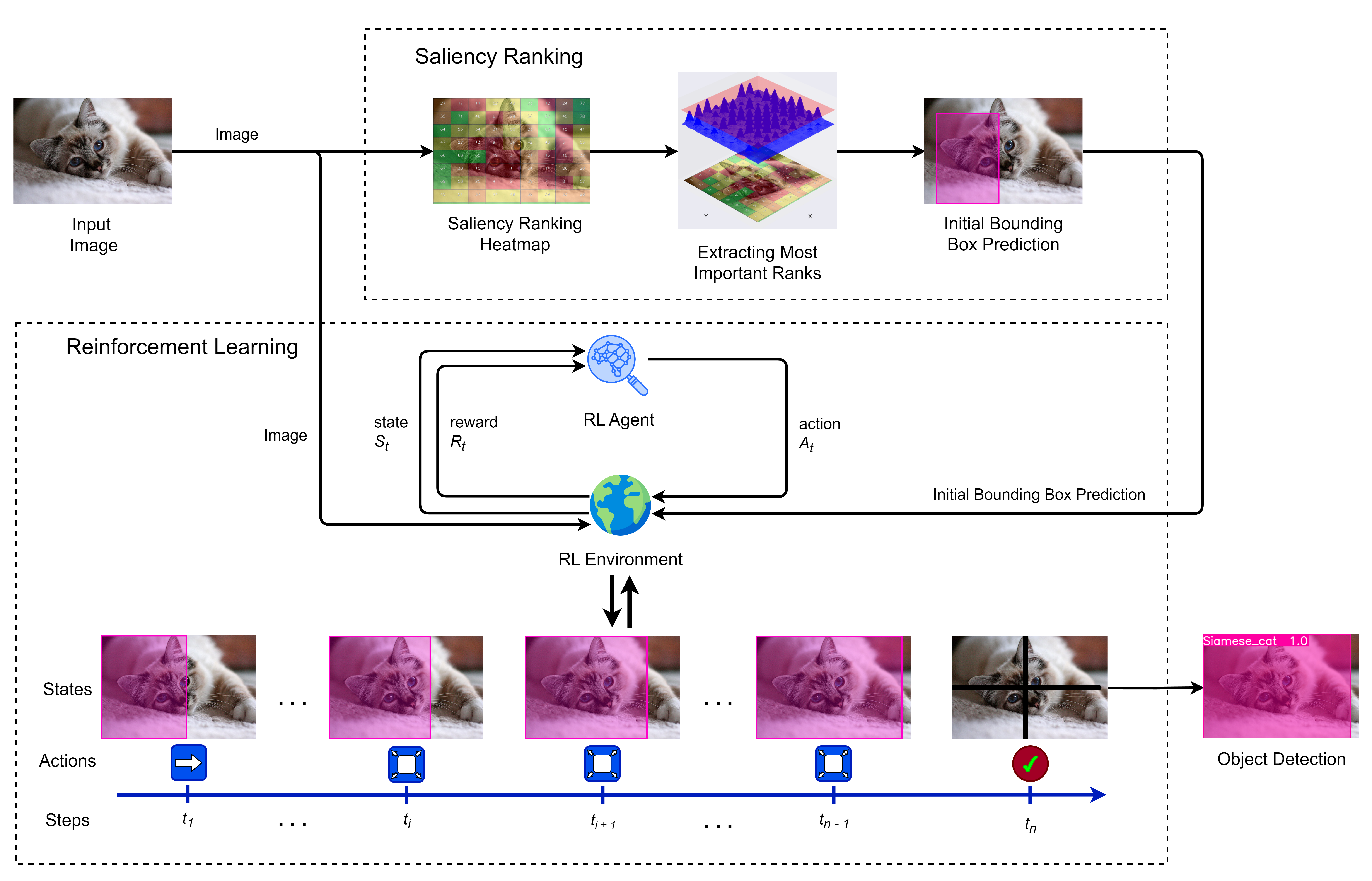}
    \caption{SaRLVision architecture.}
    \label{fig:architecture}
\end{figure}

\begin{figure}[!htbp]
    \centering
    \includegraphics[width=0.84\linewidth]{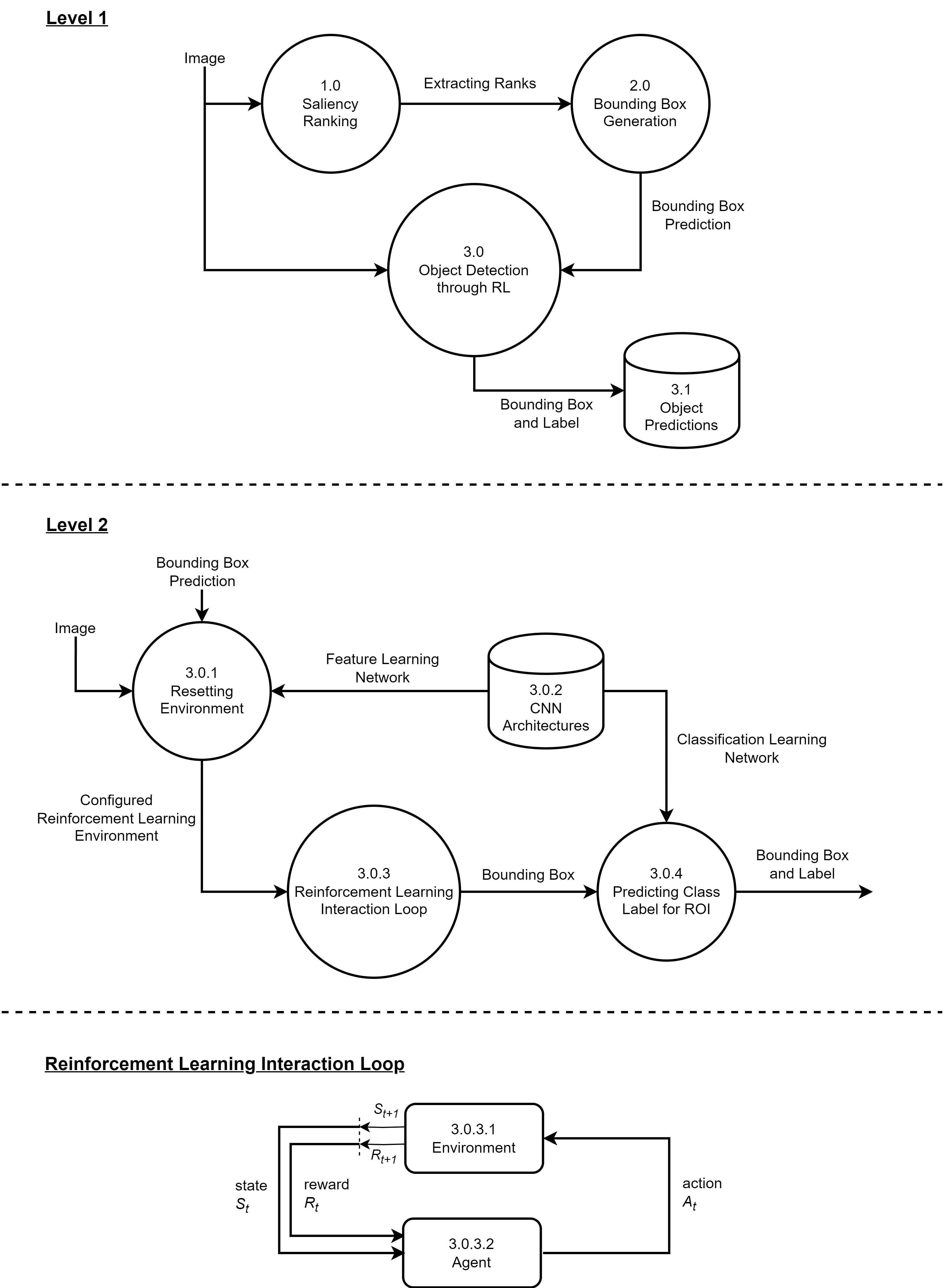}
    \caption{A data-flow diagram showcasing system pipeline.}
    \label{fig:dfd}
\end{figure}

A detailed overview of the primary components and process flow is provided by the system architecture, which is shown in the DFD in Figure \ref{fig:dfd}. This visualisation enables the understanding of the interplay between multiple modules and their sequential execution in attaining the system's goals. Additionally, Figure \ref{fig:architecture} describes the system's structure and essential features. Initially, the system proceeds to generate a saliency ranking heatmap using the input image, thus emphasising regions of interest. It then takes the most important ranks to create an initial bounding box prediction. This prediction is then fed to the RL environment, where an agent navigates through a series of time steps, repeatedly completing actions to improve the bounding box and precisely pinpointing the object within an image, while also predicting the object class label.

% N.B. to get this good diagram had to do export zoom in draw.io to 400%

\subsection{Datasets}
\label{sec:datasets}

Object detection datasets are valuable resources for training and evaluating object detection models, comprising of ground truth bounding boxes. These datasets are also equipped with labels for multiple objects inside a single image. As discussed in Chapter \ref{chp:background} and emphasised in the prevalent literature, the Pascal VOC 2007 \cite{pascal_voc_2007} and 2012 \cite{pascal_voc_2012} datasets were chosen for training the agents, since they are extensively used for training RL-based detectors, and tend to offer comprehensive annotations. Furthermore, the class-specific DQNagents underwent training for 15 epochs on the mentioned dataset, analogous to \cite{rapport, multitaskLearning, object_localization_RL}.
Additionally, the Pascal VOC 2007 test dataset was also used to evaluate the effectiveness of the trained agents, allowing for a more standardised comparison with previous approaches \cite{rapport, object_localization_RL}. The COCO dataset \cite{cocodataset}, particularly its 2017 version, stands as a renowned benchmark for object detection tasks. In experiments related to saliency ranking, the COCO 2017 dataset was employed to further evaluate the performance of the approach and ensure its ability to generalise across diverse datasets and scenarios. The Pascal VOC dataset was obtained using the \verb|torchvision|\footnote{\url{https://pytorch.org/vision/main/datasets.html}} API, while the COCO dataset was acquired through the \verb|coco|\footnote{\url{https://cocodataset.org/\#download}} API.

\section{Evaluation}%
\label{chp:eval}

This chapter evaluates the techniques described in the methodology by establishing a set of object detection metrics for evaluating the trained agents. Additionally, it delves into four experiments, culminating in a comparative analysis. The first experiment centres on identifying the optimal threshold and number of iterations for the saliency ranking algorithm within the pipeline design. The second experiment assesses the impact of varying exploration modes and the integration of the configured SaRa process. The third experiment compares the effects of employing different feature learning architectures, whilst the final experiment examines the evaluation of training variant DQN agents. Finally, the chapter presents a comparative analysis, highlighting the differences between the developed models and those existing in the literature.

\subsection{Metrics}
\label{sec:metrics}
Metrics are quantifiable measures characterised by specific mathematical properties, which are employed to assess the performance of models. In this evaluation, adhering to the Pascal VOC object detection benchmark, the IoU threshold utilised was that of 0.5. As the current problem focuses on single object detection, when multiple ground truth bounding boxes for the same class in the image were provided, the IoU for the closest ground truth was calculated, following the approach in \cite{rapport, dissertation}. Furthermore, the following metrics serve to evaluate the performance of the developed models: \newline

\noindent
True Positives (TP), False Positives (FP), and False Negatives (FN) are essential components in the calculation of Precision and Recall. Precision, a metric that signifies the proportion of relevant items retrieved by the model, is formally defined in Equation \ref{eqn:precision}.

\begin{equation}
\text{Precision} = \frac{\text{TP}}{\text{TP} + \text{FP}} = \frac{\text{IoU}(b, g) > \text{threshold}}{\text{IoU}(b, g) > \text{threshold} + \text{FP}}
\label{eqn:precision}
\end{equation}

\noindent
Recall, represented in Equation \ref{eqn:recall}, quantifies the effectiveness of retrieving relevant items by the model.

\begin{equation}
\text{Recall} = \frac{\text{TP}}{\text{TP} + \text{FN}} = \frac{\text{IoU}(b, g) > \text{threshold}}{\text{IoU}(b, g) > \text{threshold} + \text{FN}}
\label{eqn:recall}
\end{equation}

\noindent
AP, as represented by Equation \ref{eqn:average_precision}, calculates the cumulative precision for each recall value \( R_k \) in the range from 0 to \( n \), where \( n \) signifies the total number of relevant items.

\begin{equation}
\text{Average Precision (AP)} = \sum_{k=0}^{n} (R_k - R_{k-1}) \cdot P_k
\label{eqn:average_precision}
\end{equation}

\noindent
The mAP, expressed in Equation \ref{eqn:map}, computes the average precision across all classes \( N \) by summing up the individual average precision values \( \text{AP}_i \) and dividing by the total number of classes.

\begin{equation}
\text{Mean Average Precision (mAP)} = \frac{1}{N} \sum_{i=1}^{N} \text{AP}_i
\label{eqn:map}
\end{equation}

\subsection{Experiment 1 - Optimal Threshold and Number of Iterations}
\label{sec:experiment1}

The first experiment focused on determining the optimal threshold and number of iterations required to determine the saliency ranking ranks with the highest significance, for the initial bounding box generation. This experiment comprised two main components: the optimisation of the threshold, and the determination of the optimal number of iterations for this process. Each component was further divided into two distinct experiments, both of which were evaluated on a different dataset. Firstly, a comprehensive evaluation was conducted across various thresholds, ranging from 10\% to 100\% at 10\% intervals, and using both the Pascal VOC and COCO datasets. Following this, the identified thresholds were applied to sizeable subsets of the Pascal VOC 2007+2012 training dataset (approximately 12,880 images) and the COCO 2017 training dataset (approximately 78,018 images), with a maximum subset limit of 1000 images per class, selected for the COCO dataset to mitigate class imbalance. In this experiment, the average IoU between the generated bounding box and the closest ground truth object bounding box was utilised as a metric to ascertain the optimal threshold.

\begin{table*}[ht]
\centering
% \footnotesize
\begin{tabular}{|l|c|c|}\hline
\textbf{Dataset}& \textbf{COCO 2017 Train}& \textbf{Pascal VOC 2007+2012 Train}\\ \hline \hline
\textbf{Number of Images}& 78018& 12880\\ \hline
\textbf{Optimal Threshold}& 0.3 (30\%)& 0.3 (30\%)\\ \hline
\textbf{Optimal Number of Iterations}& 2& 1\\ \hline
\textbf{Average IoU}& 0.1& 0.3\\ \hline
\textbf{Number of Categories}& 80& 20\\ \hline
\end{tabular}
\caption{Findings from the optimal threshold and number of iterations experiment.}
\label{tab:threshold_results}
\end{table*}

\begin{figure}[ht]
  \centering
  \begin{subfigure}[b]{\linewidth}
    \centering
    \includegraphics[width=\linewidth]{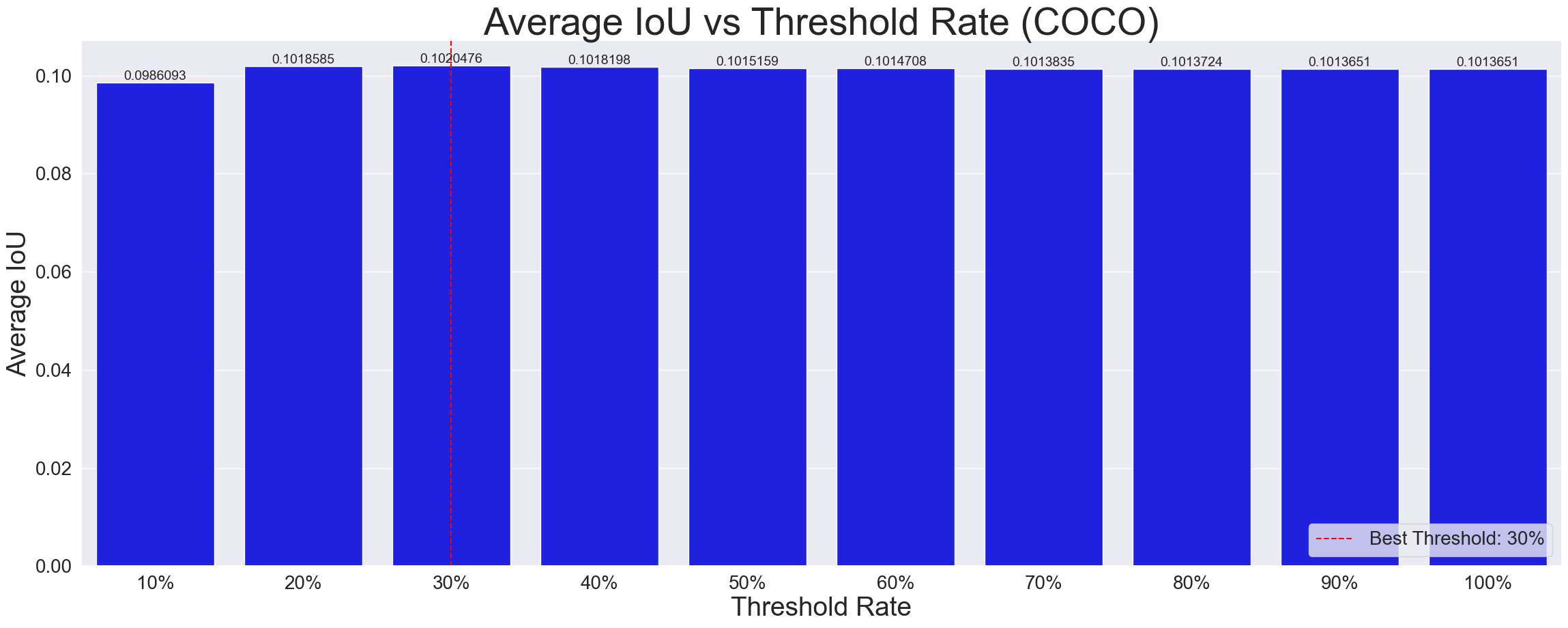}
    \caption{COCO 2017 Train}
  \end{subfigure}
  \\ [6pt]
  \begin{subfigure}[b]{\linewidth}
    \centering
    \includegraphics[width=\linewidth]{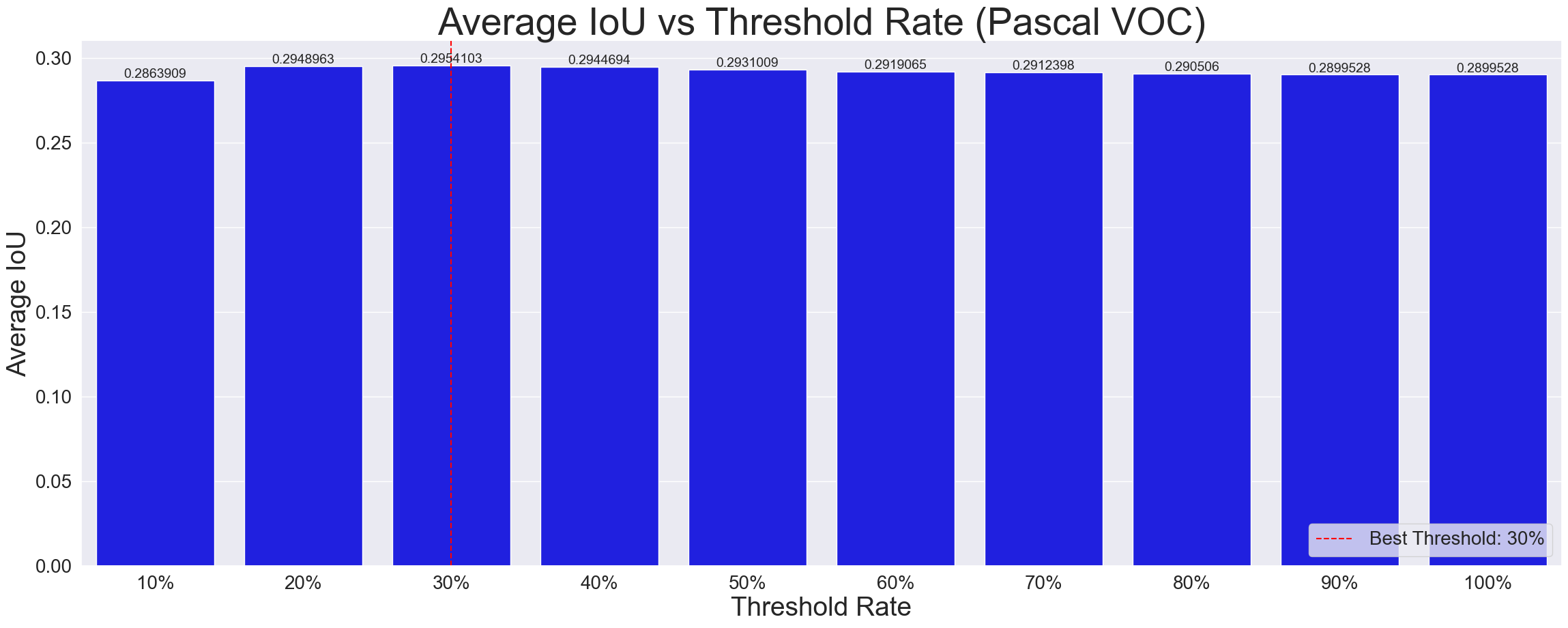}
    \caption{Pascal VOC 2007+2012 Train}
  \end{subfigure}
  \caption{Findings for determining the optimal threshold.}
  \label{fig:threshold}
\end{figure}

\begin{figure}[ht]
    \centering
    \includegraphics[width=\linewidth]{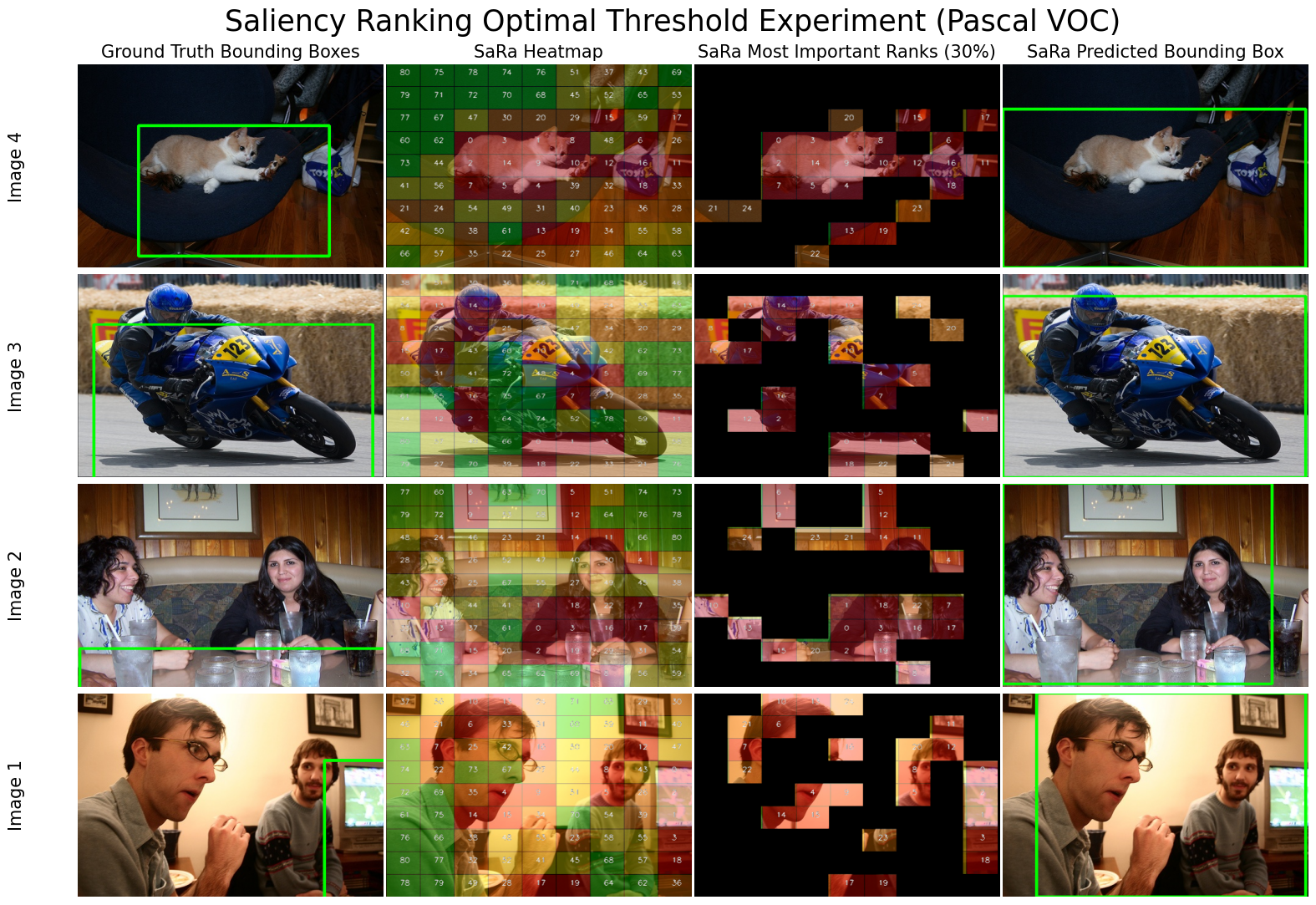}
    \caption{A selection of four images from the Pascal VOC 2007+2012 Train dataset, showcasing bounding box generation.}
    \label{fig:pascal_voc_images}
\end{figure}

\begin{figure}[ht]
  \centering
  \begin{subfigure}[b]{\linewidth}
    \centering
    \includegraphics[width=\linewidth]{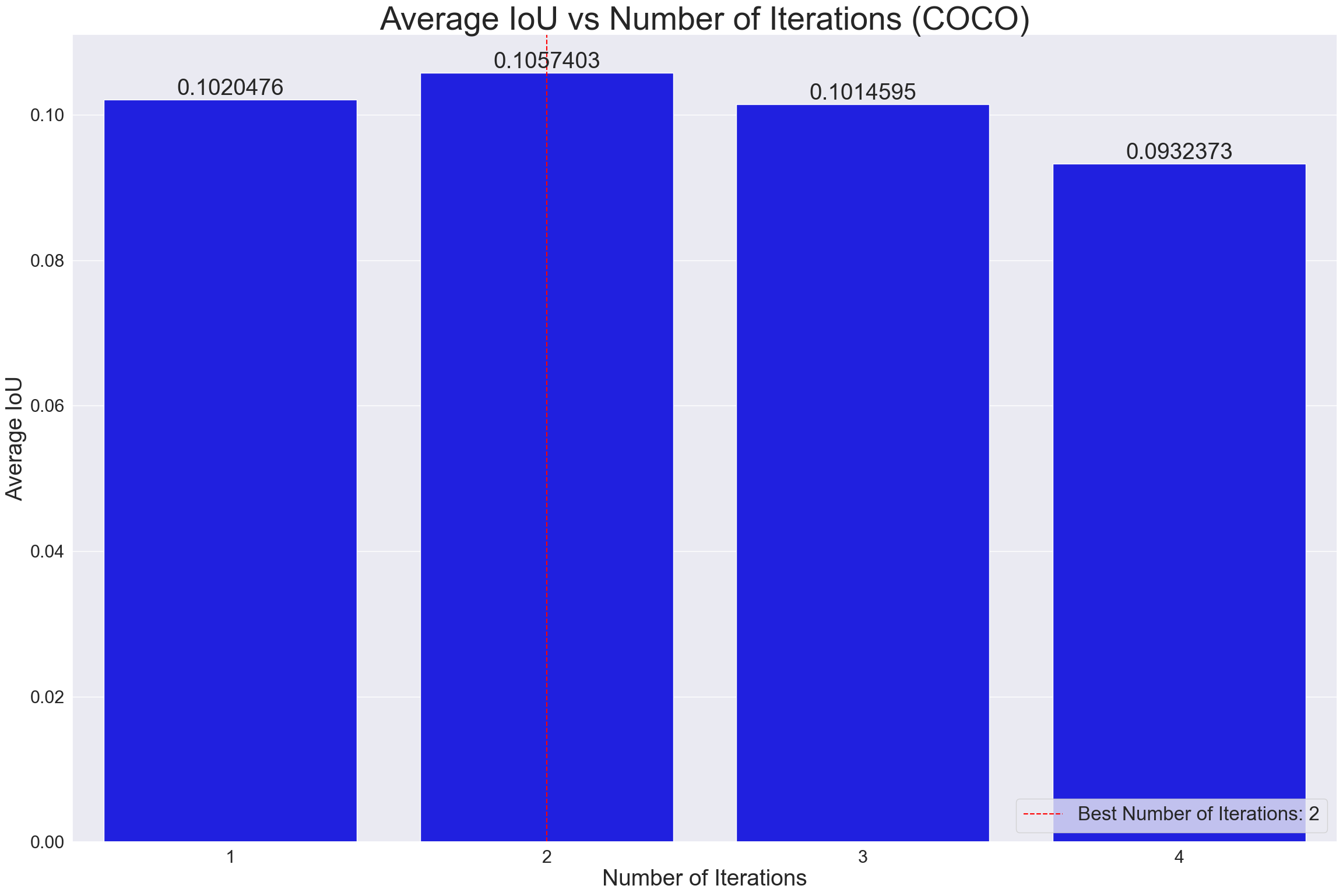}
    \caption{COCO 2017 Train}
  \end{subfigure}
  \\ [6pt]
  \begin{subfigure}[b]{\linewidth}
    \centering
    \includegraphics[width=\linewidth]{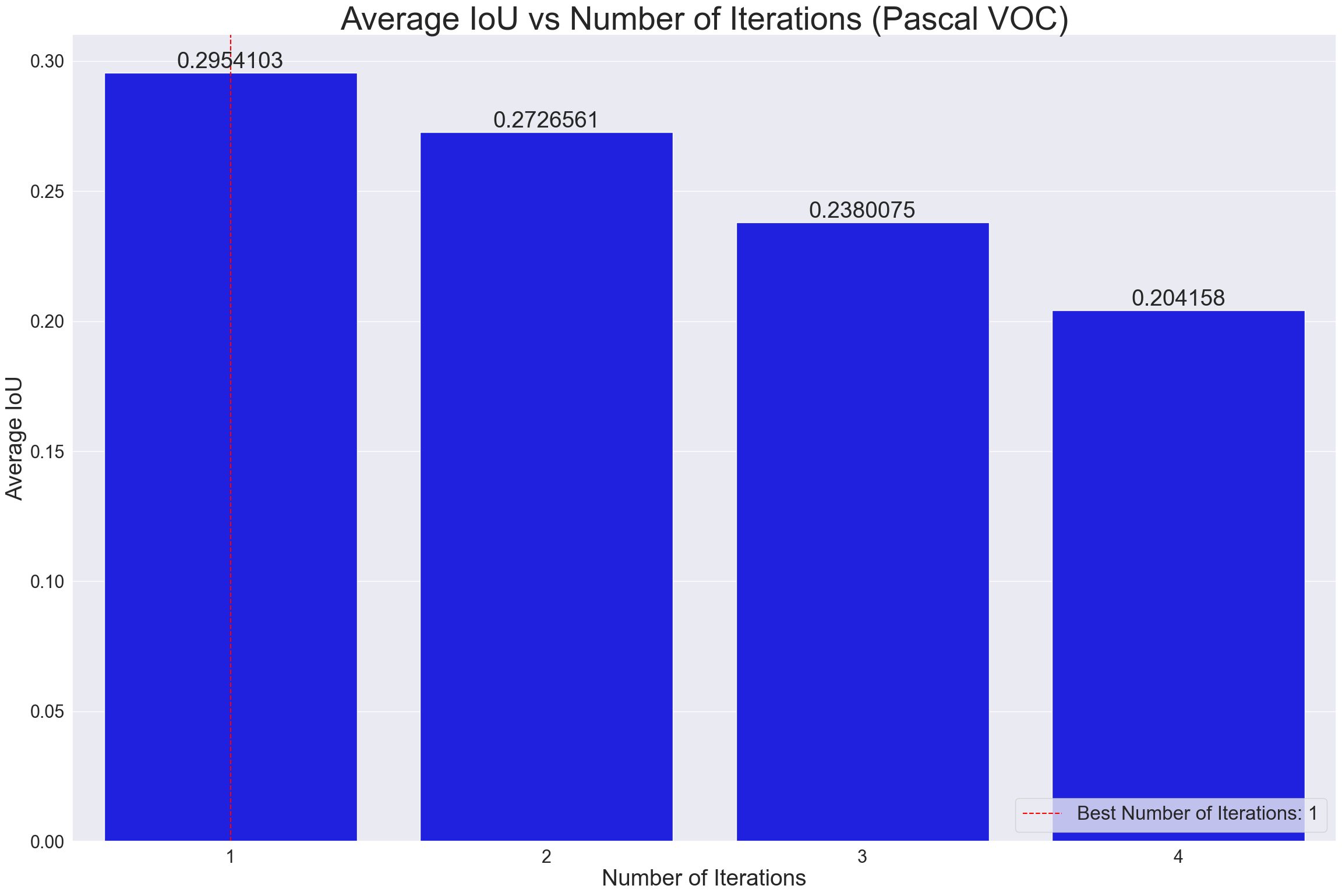}
    \caption{Pascal VOC 2007+2012 Train}
  \end{subfigure}
  \caption{Findings for determining the optimal number of iterations.}
  \label{fig:iterations}
\end{figure}

The second component involved determining the optimal number of iterations for the initial bounding box prediction process. Similar to the threshold optimisation, this experiment was conducted on both datasets, varying the number of iterations from 1 to 4 for each dataset, whilst utilising the same evaluation criteria as above. The results, presented in Table \ref{tab:threshold_results} and detailed in Figures \ref{fig:threshold} and \ref{fig:iterations}, revealed that a 30\% threshold yields optimal results, achieving an average IoU of 0.1 on the COCO dataset and 0.3 on the Pascal VOC dataset, respectively. Moreover, a minimal discrepancy was observed in the optimal number of iterations, with the evaluation on the COCO dataset suggesting 2 iterations and 1 iteration on the Pascal VOC dataset. This disparity was understandable considering the larger dataset, image size and broader categories of the COCO dataset, as illustrated in Table \ref{tab:threshold_results}.

\subsection{Experiment 2 - Exploration and Saliency Ranking}
\label{sec:experiment2}

The second experiment involved evaluating the effects of various exploration modes outlined in Chapter \ref{chp:methodology}. It explored the integration of the saliency ranking algorithm into the system pipeline following the optimal SaRa configuration determined in the previous experiment (Section \ref{sec:experiment1}). Throughout this experiment, the feature learning architecture, specifically VGG16 chosen for its smaller state size, and the standard DQN architecture remained consistent.

\begin{table*}[ht]
    \footnotesize
    \centering
    \scalebox{0.98}{
    \begin{tblr}{|l|c|c|c|c|c|c|}
        \hline
        \SetCell[r=2]{c} \textbf{Index} & \SetCell[c=5]{c} \textbf{Configuration} & & & & & \SetCell[r=2,c=1]{c} \textbf{$\text{mAP}^{\text{IoU=.50}}$} \\
        \cline{2-6}
        & \textbf{Agent} & \textbf{Exploration} & \textbf{Feature Network} & \textbf{SaRa Trained} & \textbf{SaRa Inference} & \\
        \hline\hline
        \textbf{Config 1)} & DQN & Random & VGG16 & No & No & \textbf{46.86}\\\hline
        \textbf{Config 2)} & DQN & Random & VGG16 & No & Yes & 39.23\\\hline
        \textbf{Config 3)} & DQN & Random & VGG16 & Yes & Yes & \textbf{42.12}\\\hline
        \textbf{Config 4)} & DQN & Random & VGG16 & Yes & No & 35.73\\\hline
        \textbf{Config 5)} & DQN & Guided & VGG16 & No & No & 40.10\\\hline
        \textbf{Config 6)} & DQN & Guided & VGG16 & No & Yes & 41.54\\\hline
        \textbf{Config 7)} & DQN & Guided & VGG16 & Yes & Yes & 32.67\\\hline
        \textbf{Config 8)} & DQN & Guided & VGG16 & Yes & No & 33.70\\\hline
    \end{tblr}
    }
    \caption{Exploring the effects of changing exploration modes and utilisation of saliency ranking on the Pascal VOC 2007 Test Set.}
    \label{tab:experiment2_results}
\end{table*}

The results, summarised in Table \ref{tab:experiment2_results}, indicated a significant disparity between the mAP metric, favouring random exploration over guided exploration. Notably, the most effective configuration involved random exploration without the inclusion of saliency ranking in the initial bounding box generation. This outcome suggests that restricting the agent's observability to the extracted feature vector of the observed region may have played a role. Introducing saliency ranking, which reduces the primary observed region, might have confused the agent rather than the aiding in narrowing down the region of interest.
Furthermore, additional tests were conducted to evaluate the impact of applying or disabling the initial SaRa process during inference on mAP. As shown in Table \ref{tab:experiment2_results}, the results varied. In some instances, adding the SaRa prediction process to agents not trained with it improved mAP, while in other cases, it led to poorer performance. Similarly, disabling the process for trained agents resulted in inconsistent outcomes, even after multiple tests.

\begin{figure}[!htbp]
    \centering
    \includegraphics[width=\linewidth]{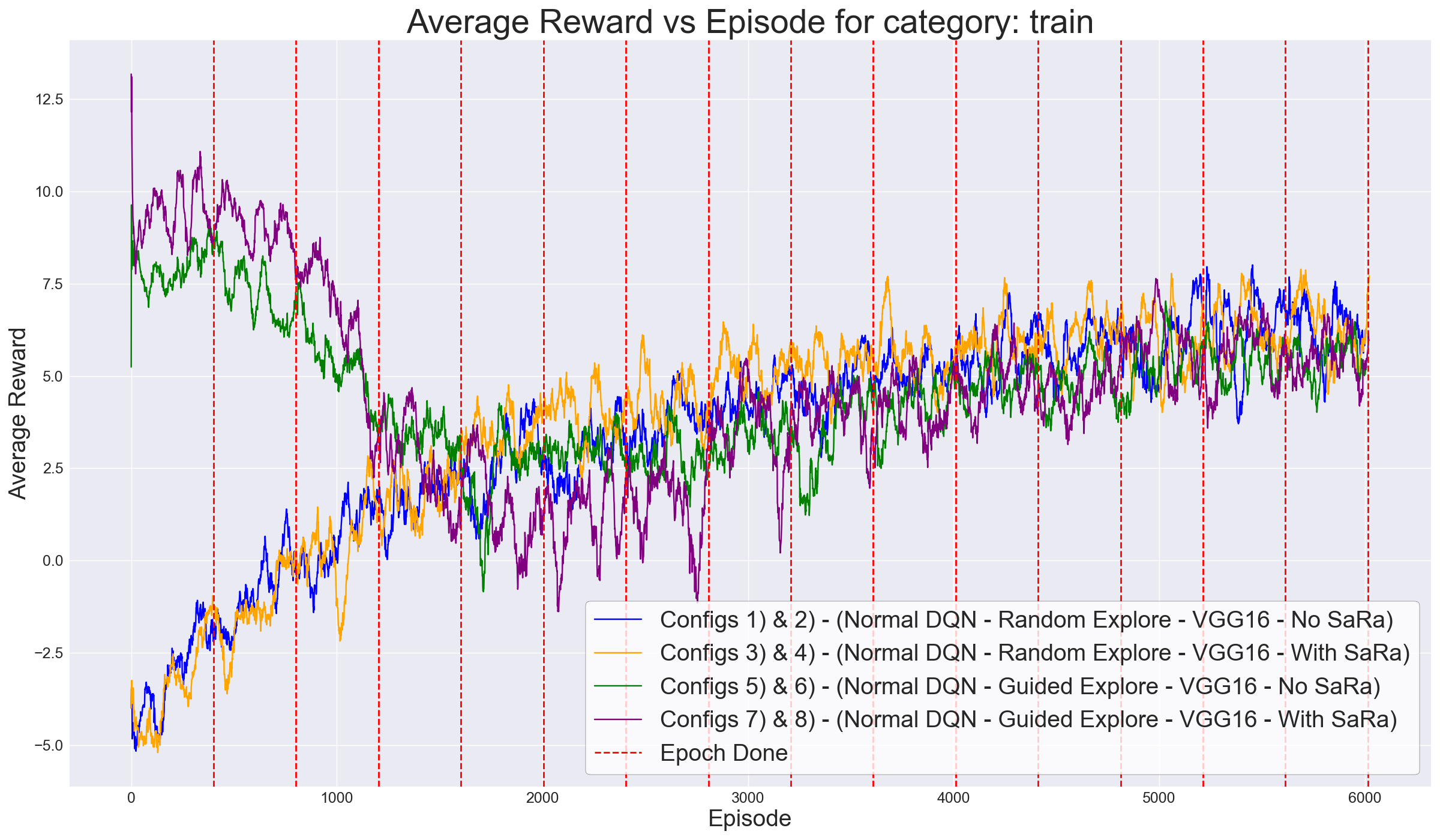}
    \caption{The average reward per episode for the 'train' category in Experiment 2. This shows the influence of employing saliency ranking while varying the exploration modes during training on the Pascal VOC 2007+2012 Train Set.}
    \label{fig:exp2_total_reward}
\end{figure}

\begin{figure}[!htbp]
    \centering
    \includegraphics[width=1\linewidth]{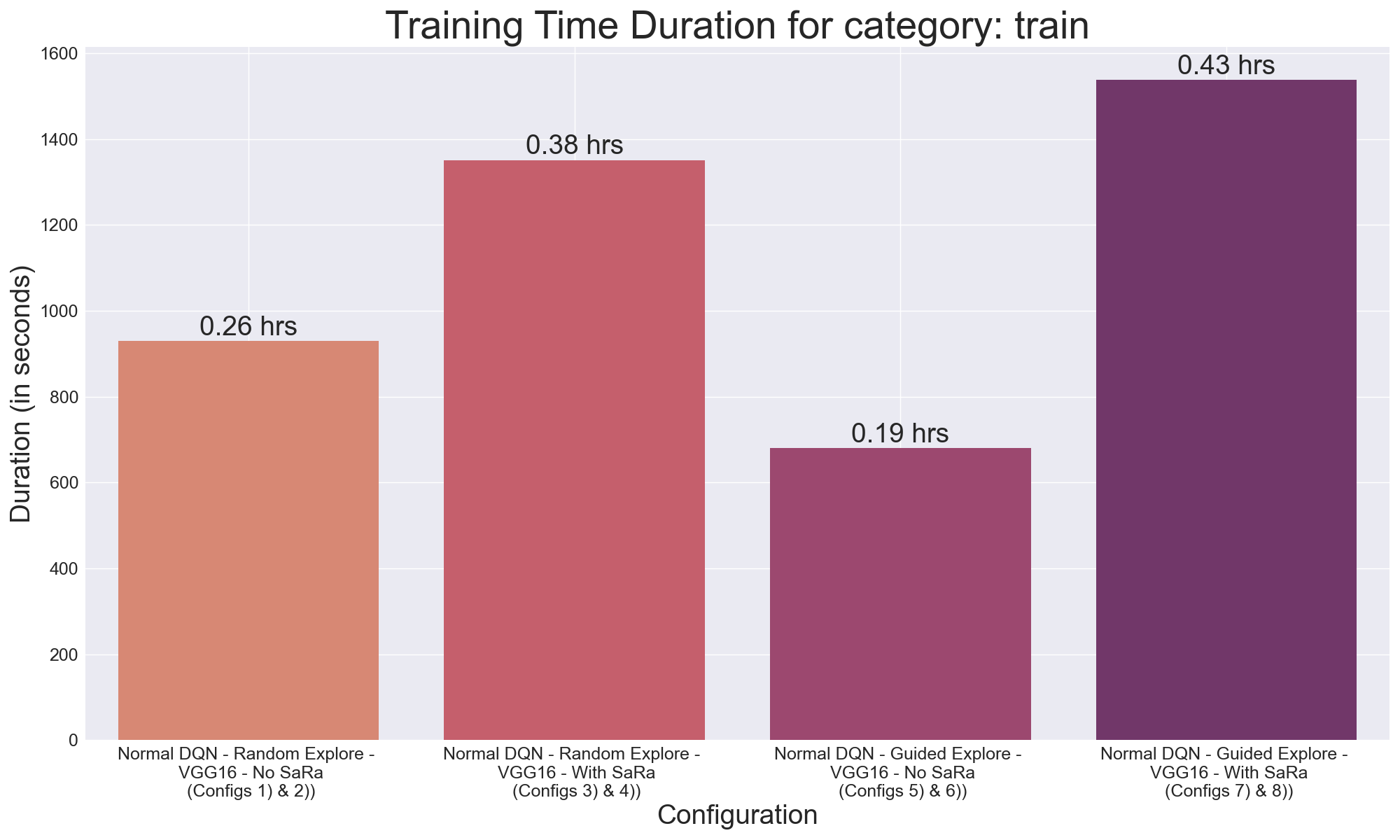}
    \caption{Training Time for Experiment 2.}
    \label{fig:exp2_training_time}
\end{figure}

Moreover, when examining the average reward across episodes during the training process, as illustrated in Figure \ref{fig:exp2_total_reward}, it is clearly evident that the reward trajectory differs between random exploration and guided exploration. In random exploration, the reward initially starts at a relatively low value and gradually increases as the agent transitions from exploration to exploitation. Conversely, in guided exploration, the reward begins at a high value and gradually decreases as the agent shifts towards exploitation. This phenomenon is attributed to the nature of guided exploration, which benefits from full observability in calculating the IoU from the ground truth bounding boxes. In contrast, when predicting the location of an object in an unseen image, such full observability is not available, thus explaining the decline in value when transitioning to exploitation. This observation can also be credited to the nature of guided exploration, which relies on selecting random actions with positive rewards. It is worth noting that both modes eventually converge to approximately the same reward level. Figure \ref{fig:exp2_total_reward} also highlights that the application of the SaRa initial prediction results in greater variation in the training period, as evidenced by the wide fluctuation in the total reward. Additionally, in line with previous studies \cite{object_localization_RL, rapport, dissertation}, it is observed that the agent transitions from exploration to exploitation approximately occurs after the third epoch for the current category.
Furthermore, it is also apparent that guided exploration leads to a faster convergence rate compared to random exploration. This finding is consistent with its directed approach of exploration, as illustrated in detail in Figure \ref{fig:exp2_training_time}. Moreover, the utilisation of SaRa leads to slower execution, attributed to its algorithmic structure featuring multiple iterative processes. Upon comparing the training time with previous studies \cite{rapport}, it is noted that there is a significant reduction in training time in the proposed approach. Whilst, \cite{rapport} reports approximately three hours of training time for each category, the designed approach reduces this to less than an hour. This improvement can be attributed to the reduction in state size, enabling the designed models to be more lightweight and faster.

\subsection{Experiment 3 - Feature Learning Architectures}
\label{sec:experiment3}

The third experiment evaluates the effects of employing diverse feature learning architectures as outlined in Chapter \ref{chp:methodology}. Building upon the insights gained from the preceding experiment in Section \ref{sec:experiment2}, which identified the top two performing streams utilising random exploration, one incorporating saliency ranking and the other without it, this experiment further investigates their effectiveness across diverse feature learning architectures. Throughout this experimentation, the random exploration mode and the standard DQN architecture were retained.

\begin{table*}[ht]
    \footnotesize
    \centering
    \scalebox{0.98}{
    \begin{tblr}{|l|c|c|c|c|c|c|}
        \hline
        \SetCell[r=2]{c} \textbf{Index} & \SetCell[c=5]{c} \textbf{Configuration} & & & & & \SetCell[r=2,c=1]{c} \textbf{$\text{mAP}^{\text{IoU=.50}}$} \\
        \cline{2-6}
        & \textbf{Agent} & \textbf{Exploration} & \textbf{Feature Network} & \textbf{SaRa Trained} & \textbf{SaRa Inference} & \\
        \hline\hline
        \textbf{Config 1)} & DQN & Random & VGG16 & No & No & \textbf{46.86}\\\hline
        \textbf{Config 3)} & DQN & Random & VGG16 & Yes & Yes & 42.12\\\hline
        \textbf{Config 9)} & DQN & Random & MobileNet & No & No & 41.42\\\hline
        \textbf{Config 10)} & DQN & Random & MobileNet & Yes & Yes & 44.71\\\hline
        \textbf{Config 11)} & DQN & Random & ResNet50 & No & No & 43.16\\\hline
        \textbf{Config 12)} & DQN & Random & ResNet50 & Yes & Yes & 35.22\\\hline
    \end{tblr}
    }
    \caption{Exploring the effects of varying different feature learning architectures on the Pascal VOC 2007 Test Set.}
    \label{tab:experiment3_results}
\end{table*}

The findings from this experiment, as depicted in Table \ref{tab:experiment3_results}, indicate that VGG16 performed the best among the chosen image feature extraction networks, followed by MobileNet and ResNet50. Moreover, while most results showed improvement compared to the previous experiment with higher mAP values, there appears to be a correlation between the state size and mAP. Smaller state sizes not only facilitates lighter-weight models but also achieve higher mAP results compared to larger state sizes, such as when using ResNet50.
Additionally, while the use of the SaRa initial bounding box prediction process generally led to decreased mAP accuracy, it is worth noting that this was not observed when employing the MobileNet feature learning network. Interestingly, employing SaRa resulted in a higher mAP score than when it was disabled. Although this could be considered an anomaly, repeated testing yielded consistent results, suggesting a potential correlation in employing SaRa in similar state sizes that achieve comparable results.

\begin{figure}[!htbp]
    \centering
    \includegraphics[width=\linewidth]{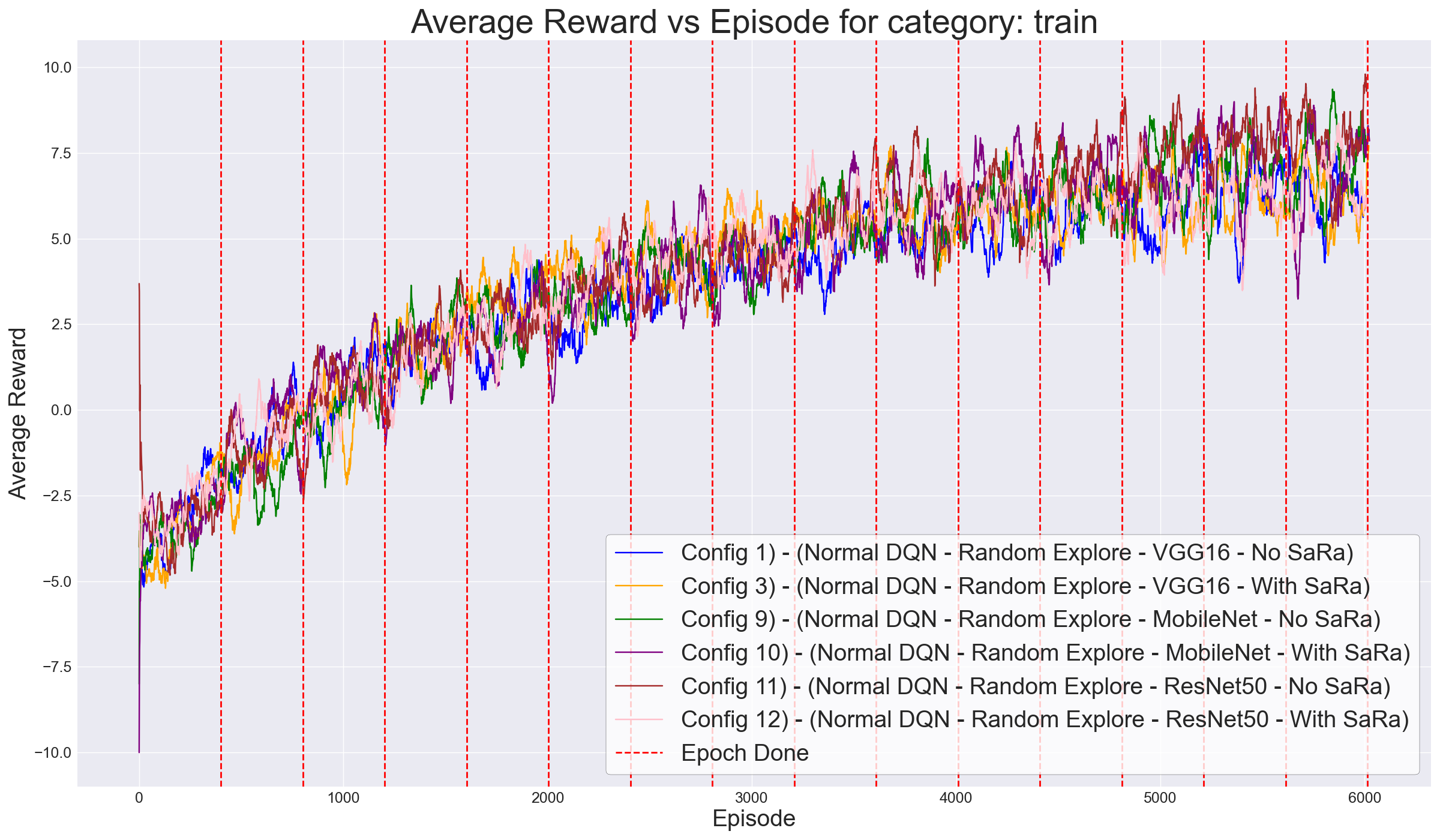}
    \caption{The average reward per episode for the 'train' category in Experiment 3. This illustrates the impact of utilising different feature learning architectures during training on the Pascal VOC 2007+2012 Train Set.}
    \label{fig:exp3_total_reward}
\end{figure}

Furthermore, when analysing the average reward over episodes during the training phase, as depicted in Figure \ref{fig:exp3_total_reward}, it becomes apparent that all configurations exhibit similar rates of convergence. This observation aligns with the findings reported in the previous experiment in Section \ref{sec:experiment2}, where the utilisation of SaRa led to increased fluctuations during training. It is evident that VGG16 demonstrates the most gradual and rapid convergence, while the MobileNet configuration experiences the least fluctuations during training, albeit achieving slightly slower convergence compared to VGG16. Proof of this is exhibited in Figure \ref{fig:exp3_total_reward}.
Conversely, the ResNet50 configuration exhibits a more linear convergence pattern than the typical curved convergence observed in such systems. Once again, this observation can be directly linked to the state size, suggesting a correlation between convergence speed, type, and fluctuations. Smaller state sizes tend to result in faster and more gradual convergence, whereas larger state sizes lead to slower, linear convergence. Remarkably, an optimal convergence, characterised by minimal fluctuations and a balanced blend of gradual and rapid convergence, may be achieved with a state size between those of the VGG16 and MobileNet networks.

\begin{figure}[!htbp]
    \centering
    \includegraphics[width=1\linewidth]{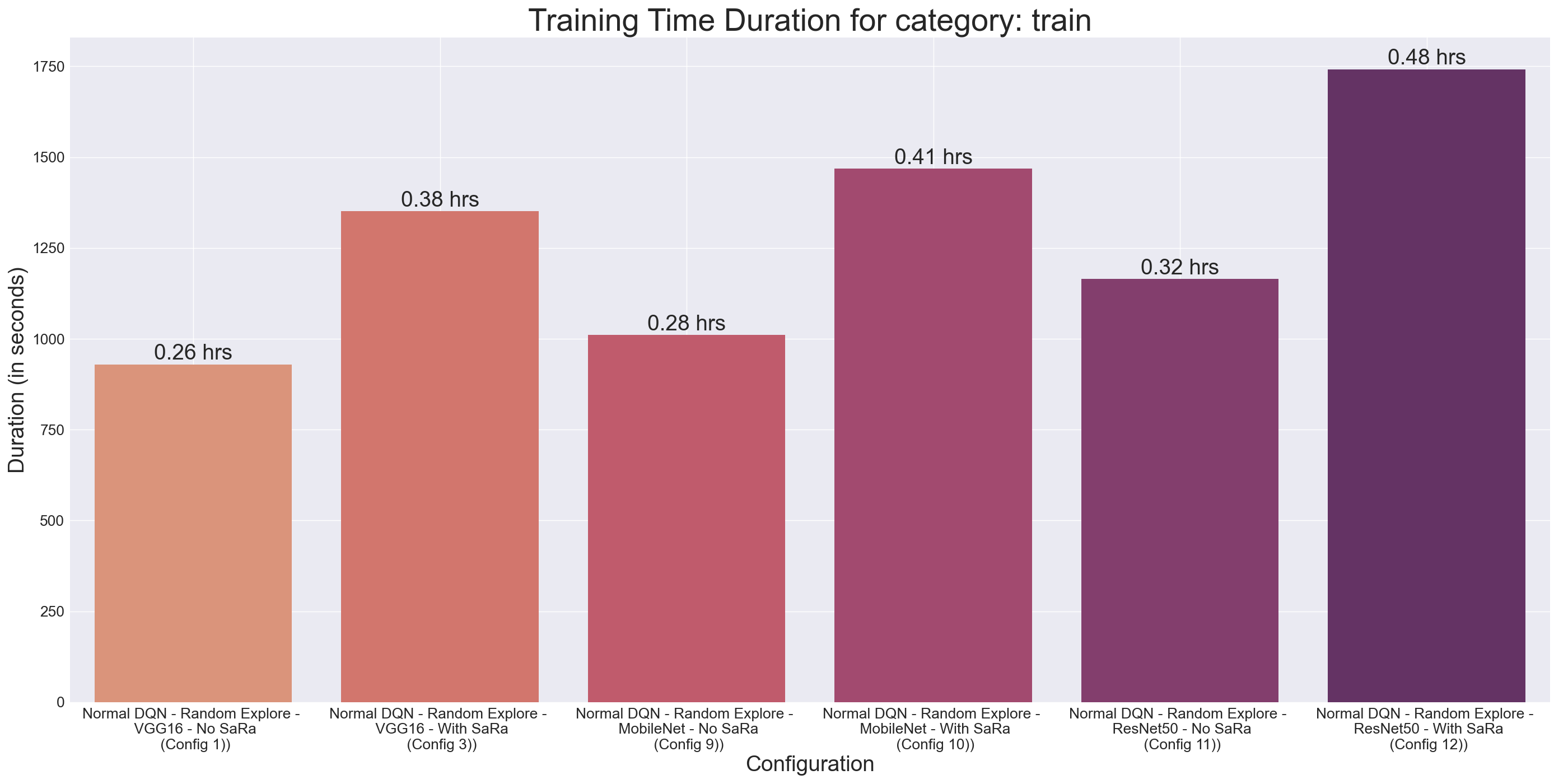}
    \caption{Training Time for Experiment 3.}
    \label{fig:exp3_training_time}
\end{figure}

Finally, it is also worth mentioning that there is a direct correlation between increasing the state size and the corresponding augmentation in execution time, as supported by Figure \ref{fig:exp3_training_time}. However, it is intriguing to observe that this augmentation is relatively marginal, resulting in only a modest disparity. Furthermore, it is discernible that the discrepancy in time overhead between employing SaRa as an initial prediction or not, is substantially smaller compared to the impact of increasing the state size. Evidently, the utilisation of the ResNet50 feature learning network yields swifter execution times in comparison to employing VGG16 with the use of SaRa, as illustrated in Figure \ref{fig:exp3_training_time}.

\subsection{Experiment 4 - DQN Agent Architectures}
\label{sec:experiment4}

The fourth experiment assessed the impacts of utilising various DQN architecture variants, as detailed in Chapter \ref{chp:methodology}. Expanding on the findings from the previous experiments outlined in Sections \ref{sec:experiment2} and \ref{sec:experiment3}, which pinpointed the optimal environment configuration involving random exploration and the exclusion of the initial SaRa prediction, this experiment delves deeper into their efficacy across different DQN architectures. Throughout this experimentation, the consistent random exploration mode, feature learning network, and absence of SaRa, were maintained.

\begin{table*}[ht]
    \footnotesize
    \centering
    \scalebox{0.91}{
    \begin{tblr}{|l|c|c|c|c|c|c|}
        \hline
        \SetCell[r=2]{c} \textbf{Index} & \SetCell[c=5]{c} \textbf{Configuration} & & & & & \SetCell[r=2,c=1]{c} \textbf{$\text{mAP}^{\text{IoU=.50}}$} \\
        \cline{2-6}
        & \textbf{Agent} & \textbf{Exploration} & \textbf{Feature Network} & \textbf{SaRa Trained} & \textbf{SaRa Inference} & \\
        \hline\hline
        \textbf{Config 1)} & DQN & Random & VGG16 & No & No & 46.86\\\hline
        \textbf{Config 13)} & DDQN & Random & VGG16 & No & No & 39.94\\\hline
        \textbf{Config 14)} & Dueling DQN & Random & VGG16 & No & No & 48.96\\\hline
        \textbf{Config 15)} & D3QN & Random & VGG16 & No & No & \textbf{51.37}\\\hline
    \end{tblr}
    }
    \caption{Exploring the effects of varying different DQN architectures on the Pascal VOC 2007 Test Set.}
    \label{tab:experiment4_results}
\end{table*}

\begin{figure}[!htbp]
    \centering
    \includegraphics[width=\linewidth]{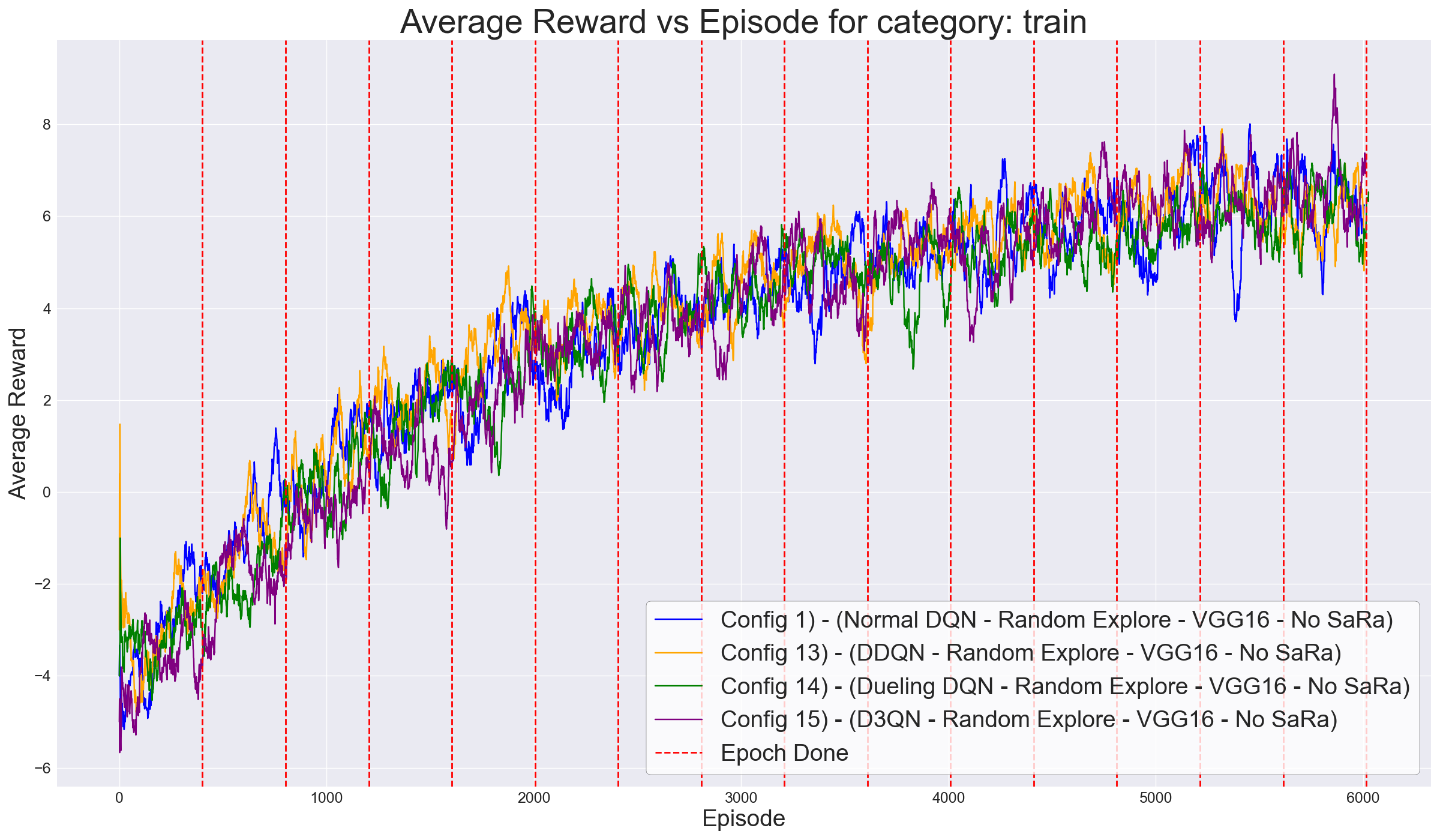}
    \caption{The average reward per episode for the 'train' category in Experiment 4. This demonstrates the effects of training with various DQN agent architectures on the Pascal VOC 2007+2012 Train Set.}
    \label{fig:exp4_total_reward}
\end{figure}

The results obtained from this experiment, as shown in Table \ref{tab:experiment4_results}, highlight D3QN as the top-performing agent, achieving a mAP of 51.37. This marks a significant improvement over the modest 46.86 mAP obtained by the standard DQN. This enhancement can largely be attributed to the architecture of the D3QN model. Interestingly, while Dueling DQN also demonstrated better performance compared to the standard DQN, DDQN yielded inferior results. However, when these architectures were combined to form D3QN, the resulting performance surpassed that of Dueling DQN, contrary to the observed pattern between standard DQN and DDQN, where the mAP decreased.
Additionally, upon analysing the average reward across episodes during the training phase, as illustrated in Figure \ref{fig:exp4_total_reward}, it is evident that all configurations demonstrate comparable rates of convergence. However, key differences between the convergence pattern of Dueling DQN akin to the standard stable DQN were recorded, as evidenced in Figure \ref{fig:exp4_total_reward}. Notably, the convergence pattern of Dueling DQN exhibits more pronounced fluctuations in average reward compared to standard DQN. These fluctuations are attributed, at least in part, to architectural disparities between the two algorithms. On the other hand, DDQN demonstrates smoother trends compared to standard DQN, which is attributed in part to its mechanism for mitigating overestimation bias inherent in traditional DQN. Furthermore, D3QN incorporates the advantages of Dueling DQN and DDQN, resulting in the most stable trend in the average reward trajectory, with reduced overestimation and fluctuations.

\begin{figure}[!htbp]
    \centering
    \includegraphics[width=1\linewidth]{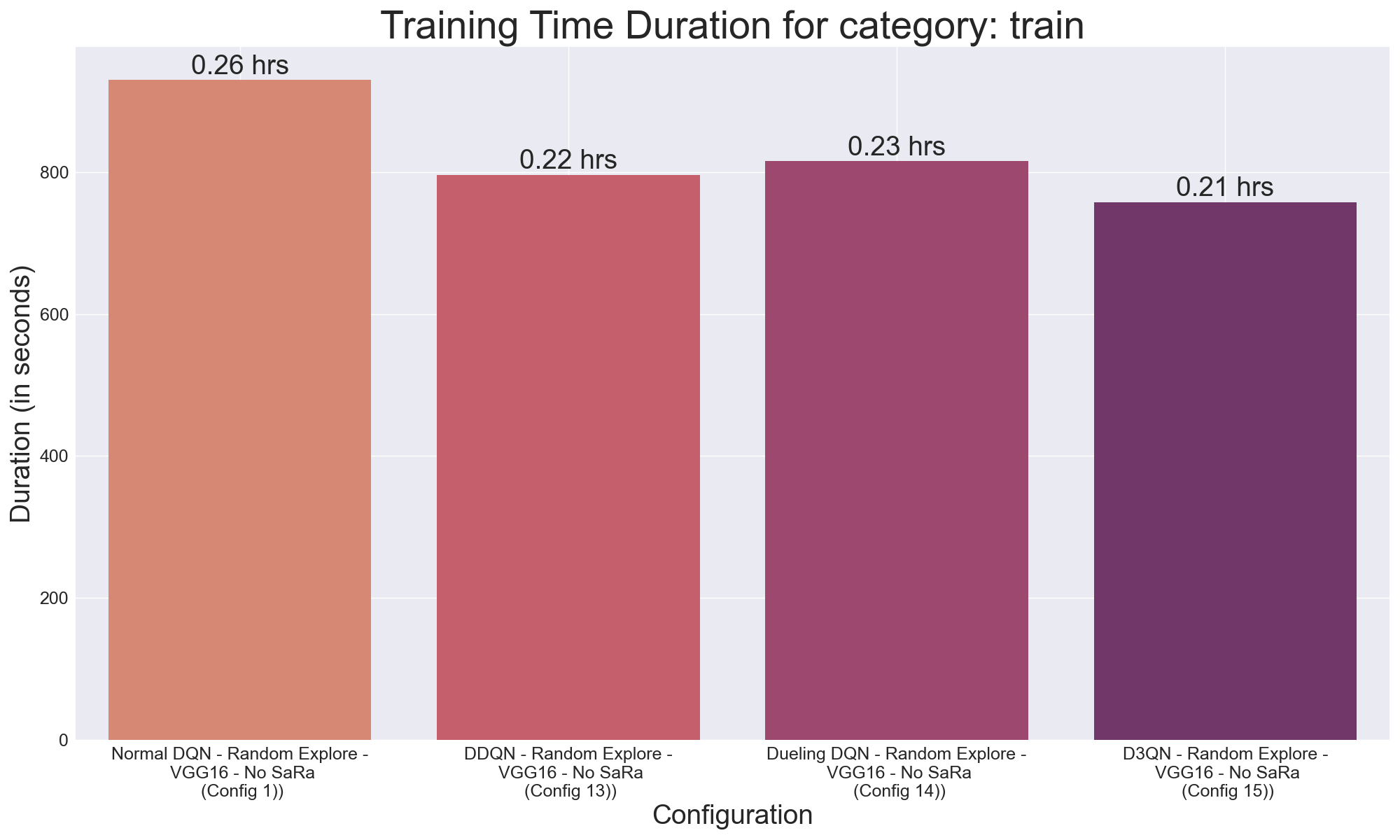}
    \caption{Training Time for Experiment 4.}
    \label{fig:exp4_training_time}
\end{figure}

It is apparent that the standard DQN agent exhibited the slowest execution time, as detailed in Figure \ref{fig:exp4_training_time}. Interestingly, although Dueling DQN's execution time was marginally slower than that of DDQN, their combination in D3QN yielded the optimal execution time. Thus, the presented agents ensure faster execution compared to standard DQN, despite the increased complexity in their algorithmic structure.

\section{Comparative Analysis}
\label{sec:analysis}
The comparison of the performance of the trained agents with the methodologies outlined in the existing literature, as discussed in Chapter \ref{chp:background}, necessitated ensuring a fair comparison. The only known single object detectors were referenced in \cite{rapport, dissertation}. Notably, the prevalent literature typically evaluated models on the Pascal VOC 2007 test set, which differs from the approach taken in \cite{dissertation}, where the 2012 version was utilised, and not all categories of this dataset were evaluated. As a result, \cite{rapport} served as the primary comparison point.
Furthermore, as demonstrated in Table \ref{tab:analysis}, it was observed that the trained agents exhibited relatively superior performance, surpassing the modest mAP score of 27.3 reported in \cite{rapport}. This improvement was particularly notable in the development of more lightweight and faster models.
Moreover, for reference purposes, Caicedo et al. \cite{object_localization_RL}, which covers multiple object detection, was also included in Table \ref{tab:analysis}. However, it is important to acknowledge that a direct comparison with this approach was not feasible due to differences in the number of object detections. In \cite{object_localization_RL}, the authors allow for multiple object proposals, precisely 200, whereas the trained agents were designed for single object detection. Thus, a fair comparison could not be achieved. Nevertheless, it is interesting to note that a majority of the trained agents exhibited performance similar to \cite{object_localization_RL} in terms of mAP scores, and in some cases, even outperformed it. Additionally, in some cases, the AP of certain categories exceeded that of \cite{object_localization_RL}, despite the limitation of only one proposal being considered, as opposed to the multiple proposals evaluated in \cite{object_localization_RL}.
Whilst the results reveal variations in the AP performance across different configurations for specific categories as depicted in Table \ref{tab:analysis}, a consistent trend emerges regarding the optimal configuration based on the mAP score. The findings indicate that employing random exploration, disabling the initial bounding box prediction process using SaRa, adopting a smaller state size with a lightweight image feature extraction network specifically the VGG16 network, and implementing more sophisticated DQN agent architectures specifically the D3QN architecture, yield the best results. These specified parameters demonstrate the optimal approach for enhancing system performance in terms of accuracy, execution time, and model size.
Finally, it is also worth mentioning that while D3QN cemented its position as the best performing agent, taking the best average precision for each category from all of the configurations, and calculating the mAP score resulted in a value of 70.6. This result continues to affirm the notion of a notably strong performance overall for the trained agents in this study.

\begin{table*}[!htbp]
\begin{center}
    \rotatebox{90}{%
        \begin{minipage}{\textheight}
            \centering
            \renewcommand{\arraystretch}{3} % Increase row height by 20%
            \scriptsize
            \scalebox{1}{
                \begin{tabular}{l|cccccccccccccccccccc|c}
                    \hline
        \textbf{Index} & \textbf{aero} & \textbf{bike} & \textbf{bird} & \textbf{boat} & \textbf{bottle} & \textbf{bus} & \textbf{car} & \textbf{cat} & \textbf{chair} & \textbf{cow} & \textbf{table} & \textbf{dog} & \textbf{horse} & \textbf{mbike} & \textbf{person} & \textbf{plant} & \textbf{sheep} & \textbf{sofa} & \textbf{train} & \textbf{tv} & \textbf{mAP} \\
        \hline \hline
        \textbf{Caicedo et al (TR) \cite{object_localization_RL}} & 57.9 & 56.7 & 38.4 & 33.0 & 17.5 & 51.1 & 52.7 & 53.0 & 17.8 & 39.1 & 47.1 & 52.2 & 58.0 & 57.0 & 45.2 & 19.3 & 42.2 & 35.5 & 54.8 & 49.0 & 43.9\\
        \textbf{Caicedo et al (AAR) \cite{object_localization_RL}} & 55.5 & 61.9 & 38.4 & 36.5 & 21.4 & 56.5 & 58.8 & 55.9 & 21.4 & 40.4 & 46.3 & 54.2 & 56.9 & 55.9 & 45.7 & 21.1 & 47.1 & 41.5 & 54.7 & 51.4 & 46.1\\
        \textbf{S. R. Ramoul \cite{rapport}} & 51.4& 30.0& 20.4& 7.0& 1.8& 43.0& 15.0& 52.7& 3.2& 24.1& 37.0& 43.7& 50.0& 45.5& 15.6& 4.4& 12.5& 39.9& 39.0& 9.4& 27.3\\ \hline
        \textbf{Config 1) (Ours)} & 76.4 & 25.7 & 64.3 & 18.3 & 4.1 & \textbf{74.6} & \textbf{67.9} & 73.1 & 4.1 & 64.3 & 34.1 & 29.8 & \textbf{82.7} & 73.9 & 38.8 & 6.6 & 68.9 & 21.3 & 52.9 & 55.7 & 46.9 \\
        \textbf{Config 2) (Ours)} & 60.2 & 14.0 & 13.5 & 60.0 & 4.3 & 70.1 & 60.1 & 68.7 & 15.3 & 12.7 & 65.1 & 18.4 & 64.3 & 21.0 & 67.8 & \textbf{65.2} & 69.2 & 20.9 & 12.5 & 1.2 & 39.2 \\
        \textbf{Config 3) (Ours)} & 70.8 & 30.3 & \textbf{72.1} & 17.7 & 19.9 & 49.3 & 63.9 & 66.9 & 6.2 & 21.6 & 33.4 & \textbf{66.0} & 70.1 & 72.6 & 22.4 & 7.2 & 44.4 & 21.5 & 72.1 & 14.1 & 42.1 \\
        \textbf{Config 4) (Ours)} & 74.0 & 65.2 & 22.6 & 9.2 & 6.7 & 35.3 & 30.0 & 34.4 & 1.2 & 22.6 & 67.9 & 25.2 & 69.2 & 32.5 & 20.8 & 58.9 & \textbf{70.0} & 23.1 & 42.9 & 2.7 & 35.7 \\
        \textbf{Config 5) (Ours)} & 65.5 & 17.8 & 67.8 & 11.3 & 2.2 & 39.1 & 64.5 & 64.8 & \textbf{65.6} & 26.5 & 66.9 & 62.0 & 74.8 & 32.0 & 22.5 & 5.0 & 16.4 & 21.6 & 71.9 & 3.8 & 40.1 \\
        \textbf{Config 6) (Ours)} & 27.6 & 26.2 & 20.0 & 12.8 & 58.6 & 72.4 & 62.1 & 59.1 & 59.1 & 14.8 & 62.5 & 60.7 & 60.7 & 18.6 & \textbf{68.8} & 1.7 & 13.8 & 59.1 & 69.7 & 2.6 & 41.5 \\
        \textbf{Config 7) (Ours)} & 74.3 & 18.8 & 21.3 & 17.7 & 1.9 & 25.7 & 66.4 & 24.3 & 4.3 & 64.8 & 71.9 & 30.1 & 30.6 & 62.3 & 22.1 & 1.8 & 12.5 & 25.3 & 73.7 & 3.9 & 32.7 \\
        \textbf{Config 8) (Ours)} & \textbf{76.8} & 16.2 & 33.2 & 18.0 & 25.2 & 31.8 & 23.2 & 64.7 & 4.9 & 22.2 & 34.5 & 39.5 & 74.1 & 23.3 & 61.5 & 6.8 & 20.2 & 25.1 & 69.8 & 2.9 & 33.7 \\
        \textbf{Config 9) (Ours)} & 56.2 & 23.6 & 45.6 & 26.5 & 3.5 & 68.6 & 67.0 & 33.7 & 4.3 & 17.5 & 68.2 & 45.1 & 76.2 & 45.2 & 20.4 & 4.8 & 66.3 & 63.9 & \textbf{76.5} & 15.4 & 41.4 \\
        \textbf{Config 10) (Ours)} & 72.6 & 62.6 & 22.0 & 18.4 & \textbf{65.1} & 33.6 & 65.2 & 44.9 & 15.4 & \textbf{65.7} & 28.2 & 63.4 & 70.1 & 30.5 & 11.2 & 10.0 & 64.9 & 18.2 & 68.0 & \textbf{64.3} & 44.7 \\
        \textbf{Config 11) (Ours)} & 74.4 & 27.6 & 35.6 & 61.5 & 14.7 & 34.2 & 64.2 & 43.9 & 4.7 & 62.8 & \textbf{73.4} & 24.8 & 81.1 & 71.8 & 16.2 & 5.3 & 22.5 & 65.4 & 76.3 & 2.9 & 43.2 \\
        \textbf{Config 12) (Ours)} & 70.6 & 19.2 & 60.2 & 30.1 & 2.4 & 37.8 & 23.6 & \textbf{75.2} & 3.2 & 61.2 & 63.3 & 37.5 & 21.6 & 22.2 & 9.1 & 4.0 & 29.6 & 63.8 & 67.8 & 1.9 & 35.2 \\
        \textbf{Config 13) (Ours)} & 64.0 & 26.5 & 20.9 & 20.7 & 7.4 & 40.8 & 48.7 & 71.5 & 4.2 & 36.3 & 25.7 & \textbf{66.0} & 80.5 & 38.1 & 25.1 & 15.0 & 63.8 & 65.2 & 73.8 & 4.4 & 39.9 \\
        \textbf{Config 14) (Ours)} & 76.4 & 62.0 & 46.9 & 62.2 & 3.1 & 69.6 & 36.6 & 66.8 & 4.0 & 29.6 & 64.1 & 23.4 & 78.1 & \textbf{75.3} & 31.2 & 62.6 & 69.6 & 33.9 & 68.7 & 14.9 & 49.0 \\
        \textbf{Config 15) (Ours)} & 76.0 & \textbf{74.2} & 67.1 & \textbf{64.7} & 4.7 & 72.7 & 64.5 & 68.7 & 3.6 & 33.7 & 23.4 & 34.0 & 77.2 & 71.5 & 64.9 & 3.2 & 23.1 & \textbf{67.5} & 73.6 & 59.3 & \textbf{51.4} \\
        \hline \hline
        \textbf{Best Category APs (Ours)} & \textbf{76.8} & \textbf{74.2} & \textbf{72.1} & \textbf{64.7} & \textbf{65.1} & \textbf{74.6} & \textbf{67.9} & \textbf{75.2} & \textbf{65.6} & \textbf{65.7} & \textbf{73.4} & \textbf{66.0} & \textbf{82.7} & \textbf{75.3} & \textbf{68.8} & \textbf{65.2} & \textbf{70.0} & \textbf{67.5} & \textbf{76.5} & \textbf{64.3} & \textbf{70.6} \\
        \hline
    \end{tabular}
            }
            \captionsetup{justification=centering} % Center caption
            \caption{Average Precision (AP) per Category and Mean Average Precision (mAP) in the Pascal VOC 2007 Test Set.}
            \label{tab:analysis}   
        \end{minipage}
    }
\end{center}
\end{table*}
\section{Conclusion}%
\label{chp:conclusion}

This study proposed innovative methodologies aimed at enhancing object detection accuracy and transparency. By integrating reinforcement learning-based visual attention techniques with saliency ranking methods, the research attempted to emulate human visual perception in object detection. Additionally, the study explored various feature learning techniques and diverse architecture variants of Deep Q-Networks, for training localisation agents via deep reinforcement learning.
The proposed approach also included real-time visualisations, allowing users to observe bounding box coordinates and action types, thereby facilitating a more intuitive understanding of algorithmic decisions.
Notably, the findings showed that although integrating saliency ranking into the system pipeline further emphasised the idea of human visual perception in object detection, in practice, this was generally impracticable. Thus, the experiments concluded that, with the omission of the saliency ranking feature, better results could be achieved. Notwithstanding, there were some specific outliers defying this conclusion, as was the case when employing the MobileNet image feature extraction network.
Moreover, the use of random exploration, generated better results than through the utilisation of guided exploration, the latter being widely mentioned in the prevalent literature. The conducted experiments also illustrated that smaller state sizes generally achieved better results in comparison to larger state sizes.
Additionally, novel trained agents, including DDQN, Dueling DQN, and D3QN, which represented uncharted avenues in the literature, were explored. Notably, Dueling DQN and D3QN exhibited significantly improved results in terms of mAP when compared to the use of standard DQN. In this study, the D3QN agent achieved the highest mAP score of 51.4, exhibiting its superior performance compared to other models.
Moreover, the results acquired demonstrated significant enhancements in single object detection methodologies employing feature networks. These improvements were achieved, in addition to obtaining comparable and even superior results to multiple object detection methods that utilise multiple object proposals, the latter not explored in the designed system. Additionally, by re-imagining the pipeline to reduce computational complexity in state size, the study enabled faster execution, greater accuracy, and the development of smaller deployable models.

\section{Future Work}
\label{chp:future_work}

Based on the findings discussed above and considering the potential of this study, several recommendations are being proposed for the advancement of research in this field. While the study addressed various enhancements in RL-based object detectors, notable areas for improvement include exploring the removal of aspect ratio constraints, and transitioning from discrete to continuous action spaces. Implementing agents capable of handling continuous action spaces could empower them to precisely control the movement of the window in any direction based on learned factors tailored to the situation, thus providing full control over image object searches.
Additionally, enhancing the current approach of discrete action spaces may entail training the Rainbow DQN \cite{rainbow_dqn_paper} agent, which amalgamates improvements from various DQN architectures, including D3QN, potentially achieving state-of-the-art performance. Furthermore, re-imagining the MDP problem may lead to better results, such as increased mAP and IoU thresholds, similar to the COCO dataset benchmark. Continuous exploration of methods to accelerate this process and avoid class-specific DQN agent training is also recommended.
Finally, additional research avenues could explore the realm of transparent and explainable AI, particularly in facilitating interpretability between input features and the resultant actions by the RL agent. While this study offers visualisations and a comprehensive action log, which are absent in the current object detectors, integrating a tool capable of elucidating the rationale behind specific actions chosen by the DQN would be crucial for future investigations. This step is essential in navigating an era where AI is increasingly prevalent, subject to legal, and ethical scrutiny.

% Need to do methodology, evaluation and continue tes tin notebook

\bibliographystyle{IEEEtran}
\bibliography{references}

\begin{thebibliography}{10}
\providecommand{\url}[1]{#1}
\csname url@samestyle\endcsname
\providecommand{\newblock}{\relax}
\providecommand{\bibinfo}[2]{#2}
\providecommand{\BIBentrySTDinterwordspacing}{\spaceskip=0pt\relax}
\providecommand{\BIBentryALTinterwordstretchfactor}{4}
\providecommand{\BIBentryALTinterwordspacing}{\spaceskip=\fontdimen2\font plus
\BIBentryALTinterwordstretchfactor\fontdimen3\font minus \fontdimen4\font\relax}
\providecommand{\BIBforeignlanguage}[2]{{%
\expandafter\ifx\csname l@#1\endcsname\relax
\typeout{** WARNING: IEEEtran.bst: No hyphenation pattern has been}%
\typeout{** loaded for the language `#1'. Using the pattern for}%
\typeout{** the default language instead.}%
\else
\language=\csname l@#1\endcsname
\fi
#2}}
\providecommand{\BIBdecl}{\relax}
\BIBdecl

\bibitem{deepRLinCV}
N.~Le, V.~S. Rathour, K.~Yamazaki, K.~Luu, and M.~Savvides, ``Deep reinforcement learning in computer vision: A comprehensive survey,'' \emph{Artif. Intell. Rev.}, vol.~55, no.~4, p. 2733–2819, 4 2022.

\bibitem{yolov10}
\BIBentryALTinterwordspacing
A.~Wang, H.~Chen, L.~Liu, K.~Chen, Z.~Lin, J.~Han, and G.~Ding, ``Yolov10: Real-time end-to-end object detection,'' 2024. [Online]. Available: \url{https://arxiv.org/abs/2405.14458}
\BIBentrySTDinterwordspacing

\bibitem{human_perception}
L.~Fei-Fei, A.~Iyer, C.~Koch, and P.~Perona, ``What do we perceive in a glance of a real-world scene?'' \emph{Journal of Vision}, vol.~7, no.~1, p.~10, 2007.

\bibitem{itti}
L.~Itti, C.~Koch, and E.~Niebur, ``A model of saliency-based visual attention for rapid scene analysis,'' \emph{IEEE Transactions on Pattern Analysis and Machine Intelligence}, vol.~20, no.~11, pp. 1254--1259, 11 1998.

\bibitem{itti_ior}
L.~Itti and C.~Koch, ``Computational modelling of visual attention,'' \emph{Nature Reviews Neuroscience}, vol.~2, no.~3, pp. 194--203, 2001.

\bibitem{object_detection_philospy}
T.~Boger and T.~Ullman, ``What is "where": Physical reasoning informs object location,'' \emph{Open Mind (Cambridge)}, vol.~7, pp. 130--140, 5 2023.

\bibitem{sara}
D.~Seychell and C.~J. Debono, ``Ranking regions of visual saliency in rgb-d content,'' in \emph{2018 International Conference on 3D Immersion (IC3D)}, 2018, pp. 1--8.

\bibitem{sara2}
D.~Seychell, ``An efficient saliency driven approach for image manipulation,'' Ph.D. dissertation, University of Malta, Valletta, Malta, 2021.

\bibitem{pySaliencyMap}
A.~Kimura, ``Saliency map implementation,'' [Online]. Available: {https://github.com/akisato-/pySaliencyMap/ }, accessed: 2017-11-1.

\bibitem{feature_learning}
M.~Jogin, Mohana, M.~S. Madhulika, G.~D. Divya, R.~K. Meghana, and S.~Apoorva, ``Feature extraction using convolution neural networks (cnn) and deep learning,'' in \emph{2018 3rd IEEE International Conference on Recent Trends in Electronics, Information \& Communication Technology (RTEICT)}, 2018, pp. 2319--2323.

\bibitem{resnet50}
K.~He, X.~Zhang, S.~Ren, and J.~Sun, ``Deep residual learning for image recognition,'' \emph{CoRR}, vol. abs/1512.03385, 2015.

\bibitem{vgg16}
K.~Simonyan and A.~Zisserman, ``{Very deep convolutional networks for large-scale image recognition},'' \emph{arXiv preprint arXiv:1409.1556}, 2014.

\bibitem{inception}
C.~Szegedy, W.~Liu, Y.~Jia, P.~Sermanet, S.~Reed, D.~Anguelov, D.~Erhan, V.~Vanhoucke, and A.~Rabinovich, ``Going deeper with convolutions,'' in \emph{2015 IEEE Conference on Computer Vision and Pattern Recognition (CVPR)}, 2015, pp. 1--9.

\bibitem{xception}
F.~Chollet, ``Xception: Deep learning with depthwise separable convolutions,'' 2017.

\bibitem{mobilenet}
A.~G. Howard, M.~Zhu, B.~Chen, D.~Kalenichenko, W.~Wang, T.~Weyand, M.~Andreetto, and H.~Adam, ``Mobilenets: Efficient convolutional neural networks for mobile vision applications,'' \emph{CoRR}, vol. abs/1704.04861, 2017.

\bibitem{efficientnet}
M.~Tan and Q.~V. Le, ``Efficientnet: Rethinking model scaling for convolutional neural networks,'' \emph{CoRR}, vol. abs/1905.11946, 2019.

\bibitem{inceptionV4}
C.~Szegedy, S.~Ioffe, V.~Vanhoucke, and A.~Alemi, ``Inception-v4, inception-resnet and the impact of residual connections on learning,'' \emph{Proceedings of the AAAI Conference on Artificial Intelligence}, vol.~31, no.~1, 2 2017.

\bibitem{imagenet1}
J.~Deng, W.~Dong, R.~Socher, L.-J. Li, K.~Li, and L.~Fei-Fei, ``Imagenet: A large-scale hierarchical image database,'' in \emph{2009 IEEE Conference on Computer Vision and Pattern Recognition}, 2009, pp. 248--255.

\bibitem{imagenet2}
O.~Russakovsky, J.~Deng, H.~Su, J.~Krause, S.~Satheesh, S.~Ma, Z.~Huang, A.~Karpathy, A.~Khosla, M.~S. Bernstein, A.~C. Berg, and L.~Fei{-}Fei, ``Imagenet large scale visual recognition challenge,'' \emph{CoRR}, vol. abs/1409.0575, 2014.

\bibitem{suttonBook}
R.~S. Sutton and A.~G. Barto, \emph{Reinforcement Learning: An Introduction}, 2nd~ed.\hskip 1em plus 0.5em minus 0.4em\relax The MIT Press, 2018.

\bibitem{rl_survey}
A.~W.~M. Leslie Pack~Kaelbling, Michael L.~Littman, ``Reinforcement learning: {A} survey,'' \emph{CoRR}, vol. cs.AI/9605103, 1996.

\bibitem{deep_rl_survey}
K.~Arulkumaran, M.~P. Deisenroth, M.~Brundage, and A.~A. Bharath, ``A brief survey of deep reinforcement learning,'' \emph{CoRR}, vol. abs/1708.05866, 2017.

\bibitem{mdp}
M.~Otterlo and M.~Wiering, ``Reinforcement learning and markov decision processes,'' \emph{Reinforcement Learning: State of the Art}, pp. 3--42, 01 2012.

\bibitem{dqn_paper}
V.~Mnih, K.~Kavukcuoglu, D.~Silver, A.~Graves, I.~Antonoglou, D.~Wierstra, and M.~A. Riedmiller, ``Playing atari with deep reinforcement learning,'' \emph{CoRR}, vol. abs/1312.5602, 2013.

\bibitem{double_dqn_paper}
H.~van Hasselt, A.~Guez, and D.~Silver, ``Deep reinforcement learning with double q-learning,'' \emph{CoRR}, vol. abs/1509.06461, 2015.

\bibitem{dueling_dqn_paper}
Z.~Wang, N.~de~Freitas, and M.~Lanctot, ``Dueling network architectures for deep reinforcement learning,'' \emph{CoRR}, vol. abs/1511.06581, 2015.

\bibitem{object_localization_RL}
J.~C. Caicedo and S.~Lazebnik, ``Active object localization with deep reinforcement learning,'' in \emph{Proceedings of the 2015 IEEE International Conference on Computer Vision (ICCV)}, ser. ICCV '15.\hskip 1em plus 0.5em minus 0.4em\relax USA: IEEE Computer Society, 2015, p. 2488–2496.

\bibitem{apprentice_learning_1}
P.~Abbeel and A.~Y. Ng, ``Apprenticeship learning via inverse reinforcement learning,'' in \emph{Proceedings of the Twenty-First International Conference on Machine Learning}, ser. ICML '04.\hskip 1em plus 0.5em minus 0.4em\relax New York, NY, USA: Association for Computing Machinery, 2004, p.~1.

\bibitem{apprentice_learning_2}
A.~Coates, P.~Abbeel, and A.~Y. Ng, ``Learning for control from multiple demonstrations,'' in \emph{Proceedings of the 25th International Conference on Machine Learning}, ser. ICML '08.\hskip 1em plus 0.5em minus 0.4em\relax New York, NY, USA: Association for Computing Machinery, 2008, p. 144–151.

\bibitem{apprentice_learning_3}
S.~Levine, C.~Finn, T.~Darrell, and P.~Abbeel, ``End-to-end training of deep visuomotor policies,'' \emph{CoRR}, vol. abs/1504.00702, 2015.

\bibitem{hierarchicaldetection}
M.~Bellver, X.~Giró-i Nieto, F.~Marques, and J.~Torres, ``Hierarchical object detection with deep reinforcement learning,'' \emph{Advances in Parallel Computing}, vol.~31, 11 2016.

\bibitem{detectionSequential}
S.~Mathe, A.~Pirinen, and C.~Sminchisescu, ``\BIBforeignlanguage{English}{Reinforcement learning for visual object detection},'' in \emph{\BIBforeignlanguage{English}{2016 IEEE Conference on Computer Vision and Pattern Recognition, CVPR 2016}}, vol. 2016-January.\hskip 1em plus 0.5em minus 0.4em\relax United States: IEEE Computer Society, 2016, pp. 2894--2902, 2016 IEEE Conference on Computer Vision and Pattern Recognition, CVPR 2016 ; Conference date: 26-06-2016 Through 01-07-2016.

\bibitem{treeStructuredRL}
Z.~Jie, X.~Liang, J.~Feng, X.~Jin, W.~F. Lu, and S.~Yan, ``Tree-structured reinforcement learning for sequential object localization,'' \emph{CoRR}, vol. abs/1703.02710, 2017.

\bibitem{multitaskLearning}
Y.~Wang, L.~Zhang, L.~Wang, and Z.~Wang, ``Multitask learning for object localization with deep reinforcement learning,'' \emph{IEEE Transactions on Cognitive and Developmental Systems}, vol.~11, no.~4, pp. 573--580, 2019.

\bibitem{rpn_2018_CVPR}
A.~Pirinen and C.~Sminchisescu, ``Deep reinforcement learning of region proposal networks for object detection,'' in \emph{Proceedings of the IEEE Conference on Computer Vision and Pattern Recognition (CVPR)}, 6 2018.

\bibitem{fastrcnn}
R.~Girshick, ``Fast r-cnn,'' \emph{CoRR}, vol. abs/1504.08083, 2015.

\bibitem{fasterrcnn}
S.~Ren, K.~He, R.~Girshick, and J.~Sun, ``Faster r-cnn: Towards real-time object detection with region proposal networks,'' in \emph{Advances in Neural Information Processing Systems}, C.~Cortes, N.~Lawrence, D.~Lee, M.~Sugiyama, and R.~Garnett, Eds., vol.~28.\hskip 1em plus 0.5em minus 0.4em\relax Curran Associates, Inc., 2015.

\bibitem{barRL}
M.~Ayle, J.~Tekli, J.~E. Zini, B.~E. Asmar, and M.~Awad, ``{BAR} - {A} reinforcement learning agent for bounding-box automated refinement,'' in \emph{The Thirty-Fourth {AAAI} Conference on Artificial Intelligence, {AAAI} 2020, The Thirty-Second Innovative Applications of Artificial Intelligence Conference, {IAAI} 2020, The Tenth {AAAI} Symposium on Educational Advances in Artificial Intelligence, {EAAI} 2020, New York, NY, USA, February 7-12, 2020}.\hskip 1em plus 0.5em minus 0.4em\relax {AAAI} Press, 2020, pp. 2561--2568.

\bibitem{linucb}
L.~Li, W.~Chu, J.~Langford, and R.~E. Schapire, ``A contextual-bandit approach to personalized news article recommendation,'' in \emph{Proceedings of the 19th International Conference on World Wide Web}, ser. WWW '10.\hskip 1em plus 0.5em minus 0.4em\relax New York, NY, USA: Association for Computing Machinery, 2010, p. 661–670.

\bibitem{uzkent2020efficient}
B.~Uzkent, C.~Yeh, and S.~Ermon, ``Efficient object detection in large images using deep reinforcement learning,'' in \emph{The IEEE Winter Conference on Applications of Computer Vision}, 2020, pp. 1824--1833.

\bibitem{ReinforceNet}
M.~Zhou, R.~Wang, C.~Xie, L.~Liu, R.~Li, F.~Wang, and D.~Li, ``Reinforcenet: A reinforcement learning embedded object detection framework with region selection network,'' \emph{Neurocomputing}, vol. 443, pp. 369--379, 2021.

\bibitem{samiei2022object}
M.~Samiei and R.~Li, ``Object detection with deep reinforcement learning,'' 2022.

\bibitem{dissertation}
M.~Otoofi, ``Object localization using deep reinforcement learning,'' \emph{Master’s thesis}, 2018, master of Science dissertation, University of Glasgow, September 8, 2018.

\bibitem{rapport}
\BIBentryALTinterwordspacing
S.~R. Ramoul, ``Rapport bibliographique: Étude de la localisation active d’objets par apprentissage par renforcement profond,'' Sorbonne Université, 11 2020, encadré par Pr. Isabelle Bloch. [Online]. Available: \url{https://github.com/rayanramoul/Active-Object-Localization-Deep-Reinforcement-Learning}
\BIBentrySTDinterwordspacing

\bibitem{pascal_voc_2007}
M.~Everingham, L.~Van~Gool, C.~K.~I. Williams, J.~Winn, and A.~Zisserman, ``The {PASCAL} {V}isual {O}bject {C}lasses {C}hallenge 2007 {(VOC2007)} {R}esults,'' http://www.pascal-network.org/challenges/VOC/voc2007/workshop/index.html.

\bibitem{pascal_voc_2012}
------, ``The {PASCAL} {V}isual {O}bject {C}lasses {C}hallenge 2012 {(VOC2012)} {R}esults,'' http://www.pascal-network.org/challenges/VOC/voc2012/workshop/index.html.

\bibitem{cocodataset}
T.~Lin, M.~Maire, S.~J. Belongie, L.~D. Bourdev, R.~B. Girshick, J.~Hays, P.~Perona, D.~Ramanan, P.~Doll{'{a} }r, and C.~L. Zitnick, ``Microsoft {COCO:} common objects in context,'' \emph{CoRR}, vol. abs/1405.0312, 2014.

\bibitem{rainbow_dqn_paper}
M.~Hessel, J.~Modayil, H.~van Hasselt, T.~Schaul, G.~Ostrovski, W.~Dabney, D.~Horgan, B.~Piot, M.~G. Azar, and D.~Silver, ``Rainbow: Combining improvements in deep reinforcement learning,'' \emph{CoRR}, vol. abs/1710.02298, 2017.

\end{thebibliography}

\end{document}